\documentclass[10pt,journal,compsoc]{IEEEtran}

\ifCLASSOPTIONcompsoc
  \usepackage[nocompress]{cite}
\else
  \usepackage{cite}
\fi

\ifCLASSINFOpdf
\else
\fi
\usepackage{amsmath}
\DeclareMathOperator\arctanh{arctanh}
\newcommand{\etal}{\textit{et al.~}}
\DeclareMathOperator*{\argmax}{arg\,max}
\DeclareMathOperator*{\argmin}{arg\,min}
\usepackage[ruled,vlined]{algorithm2e}
\usepackage{subfigure}
\usepackage{xr-hyper}
\usepackage{hyperref}
\usepackage{adjustbox}
\usepackage{multirow}
\usepackage{lipsum}
\usepackage{blindtext}
\usepackage{booktabs}
\usepackage{amsfonts}
\usepackage{amssymb}
\usepackage{mathtools}
\usepackage{color}
\usepackage{multirow}
\usepackage{array}
\usepackage{adjustbox}
\usepackage{amsmath}
\usepackage{amssymb}
\usepackage{rotating}

\usepackage{soul}
\usepackage{todonotes}

\usepackage{soul}
\usepackage{todonotes}

\begin{document}

\def\equationautorefname{Eq.}
%
\title{Universal Adversarial Audio Perturbations}
%
%
%
\author{Sajjad Abdoli,
        Luiz G. Hafemann, J\'er\^ome Rony, Ismail Ben Ayed, Patrick Cardinal and Alessandro L. Koerich
\IEEEcompsocitemizethanks{\IEEEcompsocthanksitem Authors are with \'Ecole de Technologie Sup\'erieure (\'ETS), Universit\'e du Qu\'ebec, Montreal, QC, Canada\protect\\
E-mail: sajjad.abdoli.1@ens.etsmtl.ca, luiz.gh@mailbox.org, jerome.rony.1 @etsmtl.net, \{ismail.benayed, patrick.cardinal, alessandro.koerich\} @etsmtl.ca}
\thanks{}}
\IEEEtitleabstractindextext{%
\begin{abstract}
We demonstrate the existence of universal adversarial perturbations, which can fool a family of audio classification architectures, for both targeted and untargeted attack scenarios. We propose two methods for finding such perturbations. The first method is based on an iterative, greedy approach that is well-known in computer vision: it aggregates small perturbations to the input so as to push it to the decision boundary. The second method, which is the main contribution of this work, is a novel penalty formulation, which finds targeted and untargeted universal adversarial perturbations. Differently from the greedy approach, the penalty method minimizes an appropriate objective function on a batch of samples. Therefore, it produces more successful attacks when the number of training samples is limited. Moreover, we provide a proof that the proposed penalty method theoretically converges to a solution that corresponds to universal adversarial perturbations. We also demonstrate that it is possible to provide successful attacks using the penalty method when only one sample from the target dataset is available for the attacker. Experimental results on attacking various 1D convolutional neural network architectures have shown attack success rates higher than 85.0\% and 83.1\% for targeted and untargeted attacks, respectively using the proposed penalty method.
\end{abstract}
%
}
%
\maketitle
%
\IEEEdisplaynontitleabstractindextext
%
%
\IEEEpeerreviewmaketitle
\IEEEraisesectionheading{\section{Introduction}
\label{sec:introduction}}
\IEEEPARstart{D}{eep} learning models have been achieving state-of-the-art performance in various problems, notably in image recognition \cite{zhao2019object}, natural language processing \cite{yuan2018multilingual,textstyle} and speech processing \cite{jia2018transfer,chung2018unsupervised}. However, recent studies have demonstrated that such deep models are vulnerable to adversarial attacks \cite{szegedy2013intriguing, goodfellow_explaining_2014,carlini2017towards,akhtar2018threat,grosse2019adversarial,9207309}. Adversarial examples are carefully perturbed input examples that can fool a machine learning model at test time \cite{biggio2018wild, szegedy2013intriguing}, posing security and reliability concerns for such models. The threat of such attacks has been mainly addressed for computer vision tasks \cite{akhtar2018threat, orekondy2019knockoff}. For instance, Moosavi-Dezfooli \etal\cite{Moosavi-Dezfooli_2017_CVPR} have shown the existence of {\em universal} adversarial perturbation, which, when added to an input image, causes the input to be misclassified with high probability. For these universal attacks, the generated vector is independent from the input examples.

End-to-end audio classification systems have been gaining more attention recently \cite{hannun2014deep,hoshen2015speech,oord2016wavenet,sainath2015learning}. In such systems, the input to the classifier is the audio waveform. Moreover, there have been some studies that embed traditional signal processing techniques into the layers of convolutional neural networks (CNNs) \cite{ravanelli2018speaker,zeghidour2018learning,zeghidour2018end}. For such audio classification systems, the effect of adversarial attacks is not widely addressed \cite{carlini2018audio, abdullah2020faults}. Creating attacks to threaten audio classification systems is challenging, due mainly to the signal variability in the time domain \cite{carlini2018audio}. 

In this paper, we demonstrate the existence of universal adversarial perturbations, which can fool a family of audio classification architectures, for both targeted and untargeted attack scenarios. We propose two methods for finding such perturbations. The first method is based on the greedy-approach principle proposed by Moosavi-Dezfooli \etal\cite{Moosavi-Dezfooli_2017_CVPR}, which finds the minimum perturbation that sends examples to the decision boundary. The second method, which is the main contribution of this work, is a novel penalty formulation, which finds targeted and untargeted universal adversarial perturbations. Differently from the greedy approach, the penalty method minimizes an appropriate objective function on a batch of samples. Therefore, it produces more successful attacks than the previous method when the number of training samples is limited. We also show that using this method, it is possible to provide successful attacks when only one sample from the target dataset is available to the attacker. Moreover, we provide a proof that the proposed penalty method theoretically converges to a solution that corresponds to universal adversarial perturbations. Both methods are evaluated on a family of audio classifiers based on deep models for environmental sound classification and speech recognition. The experimental results have shown that both proposed methods can attack deep models, which are used as target models, with a high success rate.

This paper is organized as follows: Section \ref{sec:adv_machine_learning} presents an overview of adversarial attacks on deep learning models. Section \ref{sec:unversal_perturbations} presents the proposed methods to craft universal audio adversarial perturbations. Section \ref{sec:exp_protocol} presents the dataset, the target models used to evaluate the proposed methods as well as the experimental results on two benchmarking datasets. are presented in Section \ref{sec:results}. The conclusion and perspective of future work is presented in the last section.
\section{Adversarial Machine Learning}
\label{sec:adv_machine_learning}
Research on adversarial attacks has attracted considerable attention recently, due to its impact on the reliability of shallow and deep learning models for computer vision tasks \cite{carlini2017towards}. For a given example $\mathbf x$, an attack can find a small perturbation $ \delta$, often imperceptible to a human observer, so that an example $\tilde{\mathbf x} = \mathbf x + \delta$ is misclassified by a machine learning model \cite{szegedy2013intriguing,peck2017lower,shafahi2018are}. The attacker's goal may cover a wide range of threats like \textit{privacy} violation, \textit{availability} violation and \textit{integrity} violation \cite{biggio2018wild}. Moreover, the attacker's goal may be \textit{specific} (targeted), with inputs misclassified as a specific class or \textit{generic} (untargeted), where the attacker simply wants to have an example classified differently from its true class \cite{biggio2018wild}. Attacks generated by targeting a specific classifier are often \emph{transferable} to other classifiers (i.e. also induce misclassification on them) \cite{fawzi2018adversarialtrans}, even if they have different architectures and do not use the same training set \cite{szegedy2013intriguing}. It has been also shown that such attacks can be applied in the physical world too. For instance, Kurakin \etal\cite{kurakin_adversarial_2017} showed that printed adversarial examples were still misclassified after being captured by a camera; Athalye \etal\cite{athalye_synthesizing_2018} presented a method to generate adversarial examples that are robust under translation and scale transformations, showing that such attacks induce misclassification even when the examples correspond to different viewpoints.

Research on adversarial attacks on audio classification systems is quite recent. Some studies focused mainly on providing inaudible and hidden targeted attacks on the systems. In such attacks, new audio examples are synthesized, instead of adding perturbations to actual inputs \cite{carlini2016hidden,zhang2017dolphinattack}. Other works focus on untargeted attacks on speech and music classification systems \cite{gong2017crafting,kereliuk2015deep}. Du \etal\cite{du2019sirenattack} proposed a method based on particle swarm optimization for targeted and untargeted attacks. They evaluated their attacks on a range of applications such as speech command recognition, speaker recognition, sound event detection and music genre classification. Alzantot \etal\cite{alzantot2018did} proposed a similar approach that uses a genetic algorithm for crafting targeted black-box attacks on a speech command classification system \cite{sainath2015convolutional}, which achieved 87\% of success. Carlini \etal\cite{carlini2018audio} proposed a targeted attack on the speech processing system DeepSpeech \cite{hannun2014deep}, which is a state-of-the-art speech-to-text transcription neural network. A penalty method is used in such attacks, which achieved 100\% of success.

Most studies consider that the attacker can directly manipulate the input of the classifiers. Crafting audio attacks that can work on the physical world (i.e. played over-the-air) presents challenges such as being robust to background noise and reverberations. Yakura \etal\cite{yakura2018robust} showed that over-the-air attacks are possible, at least for a dataset of short phrases. Qin \etal\cite{qin2019imperceptible} also reported successful adversarial attacks on automatic speech recognition systems, which remain effective after applying realistic simulated environmental distortions.

Universal adversarial perturbations are effective type of attacks \cite{Moosavi-Dezfooli_2017_CVPR} because the additive noise is independent of the input examples, but once added to any example, it may fool a deep model to misclassify such an example. Moosavi-Dezfooli \etal\cite{Moosavi-Dezfooli_2017_CVPR} proposed a greedy algorithm to provide such perturbations for untargeted attacks to images. The perturbation is generated by aggregating atomic perturbation vectors, which send the data points to the decision boundary of the classifier. Recently, Moosavi-Dezfooli \etal\cite{moosavi-dezfooli2018robustness} provided a formal relationship between the geometry of the decision boundary and robustness to universal perturbations. They have also shown the strong vulnerability of state-of-the-art deep models to universal perturbations. Metzen \etal\cite{metzen2017universal} generalized this idea to provide attacks against semantic image segmentation models. Recently, Behjati \etal\cite{behjati2019universal} generalized this idea to attack a text classifier in both targeted and non-targeted scenarios. In a different approach, Hayes \etal\cite{hayes2018learning} proposed a generative model for providing such universal perturbations. Recently, Neekhara \etal\cite{Neekhara_2019} were also inspired by this idea to generate universal adversarial perturbations which cause mis-transcription of audio signals of automatic transcription systems. They reported success rates up to 88.24\% on the hold-out set of the pre-trained Mozilla DeepSpeech model \cite{hannun2014deep}. Nonetheless, their method is designed only for untargeted attacks. For audio classification systems, the impact of such perturbations can be very strong if these perturbations can be played over the air even without knowing what the test examples would look like. 

\textcolor{black}{In this work we show the existence of UAPs for attacking several audio processing models in both targeted and also untargeted scenarios. The previous studies on UAPs for attacking audio processing systems, however, have focused only on untargeted attacking scenarios. We show that the iterative method proposed by Moosavi-Dezfooli \cite{Moosavi-Dezfooli_2017_CVPR} can be generalized for attacking audio models. As a part of this algorithm, we show that the decoupled direction and norm (DDN) attack \cite{rony2018decoupling}, which was originally proposed for targeting image processing models, can also be used in an iterative method for targeting audio models. Besides, we also propose a novel penalty optimization formulation for generating UAPs and we address new challenges such as generating perturbations when a single audio example is available to the attacker, as well as the transferability of UAPs in both targeted and untargeted scenarios. It is shown that the proposed methods generate UAPs, which generalize well across data points and they are to some extent model-agnostic (doubly universal) where UAPs are generalizable across different models specially for the untargeted attacking scenario.}
\section{Universal Adversarial Audio Perturbations}
\label{sec:unversal_perturbations}
In this section, we formalize the problem of crafting universal audio adversarial perturbations and propose two methods for finding such perturbations. The first method is based on the greedy-approach principle proposed by Moosavi-Dezfooli \etal\cite{Moosavi-Dezfooli_2017_CVPR} which finds the minimum perturbation that sends examples to the decision boundary of the classifier or inside the boundary of the target class for untargeted and targeted perturbations, respectively. The second method, which is the main contribution of this work, is a penalty formulation, which finds a universal perturbation vector that minimizes an objective function.

Let $\mu$ be the distribution of audio samples in \(\mathbb{R}^{d}\) and \(\hat{k}(\mathbf x)= \argmax_y \mathbb{P}\left(y | \mathbf x, \theta\right)\) be a classifier that predicts the class of the audio sample $\mathbf x$, where $y$ is the predicted label of $\mathbf x$ and $\theta$ denotes the parameters of the classifier. Our goal is to find a vector ${\mathbf v}$ that, once added to the audio samples can fool the classifier for most of the samples. This vector is called universal as it is a fixed perturbation that is independent of the audio samples and, therefore, it can be added to any sample in order to fool a classifier. The problem can be defined such that \(\hat{k}({\mathbf x}+ {\mathbf v}) \neq \hat{k}({\mathbf x})\) for a untargeted attack and, for a targeted attack, \(\hat{k}({\mathbf x}+ {\mathbf v}) = y_{t}\), where $y_t$ denotes the target class. In this context, the universal perturbation is a vector with a sufficiently small \(\ell_{p}\) norm, where \(p \in[1, \infty)\), which satisfies two constraints \cite{Moosavi-Dezfooli_2017_CVPR}: $\| {\mathbf v}\|_{p} \leq \xi$ and $\mathbb{P}_{{\mathbf x}\sim \mu}(\hat{k}({\mathbf x}+ {\mathbf v}) \neq \hat{k}({\mathbf x})) \geq 1-\delta$, where \(\xi\) controls the magnitude of the perturbation and \(\delta\) controls the desired fooling rate. For a targeted attack, the second constraint is defined as \(\mathbb{P}_{{\mathbf x}\sim \mu}(\hat{k}({\mathbf x}+ {\mathbf v}) = y_{t}) \geq 1-\delta\).
\subsection{Iterative Greedy Algorithm}
\label{sec:Iterative}
Let \(X=\left\{{\mathbf x}_{1}, \dots, {\mathbf x}_{m}\right\}\) be a set of $m$ audio files sampled from the distribution \(\mu\). The greedy algorithm proposed by Moosavi-Dezfooli \etal\cite{Moosavi-Dezfooli_2017_CVPR} gradually crafts adversarial perturbations in an iterative manner. For untargeted attacks, at each iteration, the algorithm finds the minimal perturbation \(\Delta  {\mathbf v}_{i}\) that pushes an example \({\mathbf x}_{i}\) to the decision boundary, and adds the current perturbation to the universal perturbation. In this study, a targeted version of the algorithm is also proposed such that the universal perturbation added to the example must push it toward the decision boundary of the target class. In more details, at each iteration of the algorithm, if the universal perturbation makes the model misclassify the example, the algorithm ignores it, otherwise, an extra \(\Delta{\mathbf v}_{i}\) is found and aggregated to the universal perturbation by solving the minimization problem with the following constraints for untargeted and targeted attacks respectively:
\begin{equation}
\begin{aligned}
\Delta  {\mathbf v}_{i} \leftarrow \argmin_{{\mathbf r}}\|{\mathbf r}\|_{2} & \\
\text{s.t.}\quad &\hat{k}\left({\mathbf x}_{i}+ {\mathbf v}+{\mathbf r}\right) \neq \hat{k}\left({\mathbf x}_{i}\right),\\
\text{or}\quad &\hat{k}\left({\mathbf x}_{i}+ {\mathbf  {\mathbf v}}+{\mathbf r}\right) = y_{t}.
\end{aligned}
\end{equation}
\noindent In order to find \(\Delta{\mathbf v}_{i}\) for each sample of the dataset, any attack that provides perturbation that misclassifies the sample, such as Carlini and Wanger $\ell_2$ attack \cite{carlini2017towards} or DDN attack \cite{rony2018decoupling}, can be used. Moosavi-Dezfooli \etal\cite{moosavi2016deepfool} used Deepfool to find such a vector. 

In order to satisfy the first constraint (\(||{\mathbf v}\|_{p} \leq \xi\)), the universal perturbation is projected on the $\ell_{p}$ ball of radius \(\xi \) and centered at 0. The projection function $\mathcal{P}_{p,\xi}$ is formulated as:
\begin{equation}
\mathcal{P}_{p, \xi}( {\mathbf v})= \argmin_{ {\mathbf v}^{\prime}}\left\| {\mathbf v}- {\mathbf v}^{\prime}\right\|_{2} \quad\text{s.t.}\quad\left\|{\mathbf v}^{\prime}\right\|_{p} \leq \xi.
\label{eq:proj}
\end{equation}
\noindent The termination criteria for the algorithm is defined such that the Attack Success Rate (ASR) on the perturbed training set exceeds a threshold \(1-\delta\). In this protocol, the algorithm stops for untargeted perturbations when:
\begin{equation}
\operatorname{ASR}\left(X,  {\mathbf v}\right) :=\frac{1}{m} \sum_{i=1}^{m}
\operatorname{\bf 1}\{ \hat{k}\left({\mathbf x}_{i}+ {\mathbf v}\right) \neq \hat{k}\left({\mathbf x}_{i}\right)\} \geq 1-\delta,
\label{eq:unt}
\end{equation}
\noindent where $\operatorname{\bf 1}\{\cdot\}$ is the true-or-false indicator function. For a targeted attack, we replace inequality $\hat{k}\left({\mathbf x}_{i}+ {\mathbf v}\right) \neq \hat{k}({\mathbf x}_{i})$ by $\hat{k}\left({\mathbf x}_{i}+ {\mathbf v}\right) =y_{t}$ in Eq.~\eqref{eq:unt}. The problem with iterative Greedy formulation is that the constraint is only defined on universal perturbation ($\|{\mathbf v}\|_{p} \leq \xi$). Therefore, the summation of the universal perturbation with the data points results in an audio signal that is out of a specific range such as $[0, 1]$, for almost all audio samples, even by selecting a small value for $\xi$. In order to solve this problem, we may clip the value of each resulting data point to a valid range.
\subsection{Penalty Method}
The proposed penalty method minimizes an appropriate objective function on a batch of samples from a dataset for finding universal adversarial perturbations. In the case of noise perception in audio systems, the level of noise perception can be measured using a realistic metric such as the sound pressure level (SPL). Therefore, the SPL is used instead of the \(\ell_{p}\) norm. In this paper, such a measure is used in one of the objective functions of the optimization problem, where one of the goals is to minimize the SPL of the perturbation, which is measured in decibel (dB) \cite{carlini2018audio}. The problem of crafting a perturbation in a targeted attack can be reformulated as the following constrained optimization problem:
\begin{equation}
\label{org:prob}
\begin{aligned}
\text{minimize} \quad & \text{SPL}( {\mathbf v}) \\
\text{s.t.} \quad & y_{t} = \argmax_{y} \mathbb{P}\left(y | {\mathbf x}_{i}+ {\mathbf v}, \theta\right)\\
\text{and} \quad & {0 \leq {\mathbf x}_{i}+ {\mathbf v} \leq 1 \quad \forall i}\\
\end{aligned}
\end{equation}
\noindent For untargeted attacks we use $y_{l} \neq \argmax_{y} \mathbb{P}\left(y | {\mathbf x}_{i}+ {\mathbf v}, \theta\right)$, where $y_l$ is the legitimate class.

Different from the iterative Greedy formulation, in the proposed penalty method the second constraint is defined on the summation of the data points and universal perturbation $({\mathbf x}_{i}+{\mathbf v})$ to keep the perturbed example in a valid range. As the constraint of the DDN attack, which is used in the iterative greedy algorithm, is that the data must be in range $[0, 1]$, for a fair comparison between methods, we impose the same constraint on the penalty method. This box constraint should be valid for all audio samples.

In Eq.~\eqref{org:prob}, the pressure level of an audio waveform (noise) can be computed as:
\begin{equation}
\text{SPL}( {\mathbf v})= 20\log_{10} P( {\mathbf v}),
\end{equation}
\noindent where $P( {\mathbf v})$ is the root mean square (RMS) of the perturbation signal $\mathbf {v}$ of length \(N\), which is given by: 
\begin{equation}\label{eq:power}
{P}( {\mathbf v})=\sqrt{\frac{1}{N} \sum_{n=1}^{N}  v_{n}^{2}},
\end{equation}
\noindent where $v_{n}$ denotes the $n$-th component of the array ${\mathbf v}$.

The optimization problem introduced in Eq.~\eqref{org:prob} can be solved by a gradient-based algorithm, which however does not enforce the box constraint. Therefore, we need to introduce a new parameter \({\mathbf w}\), which is defined in Eq.~\eqref{eq:trans} to ensure that the box constraint is satisfied.

This variable change is inspired by the $\ell_2$ attack of Carlini and Wagner \cite{carlini2017towards}. \
\begin{equation}
\label{eq:trans}
{\mathbf w}_{i}=\frac{1}{2}(\tanh ({\mathbf x}_{i}^\prime+{\mathbf v^\prime})+1),
\end{equation}
where ${\mathbf x}_{i}^{\prime}=\arctanh((2{\mathbf x}_{i}-1)*(1-\epsilon))$ and $\mathbf v^{\prime}=\arctanh$ $((2\mathbf v-1)*(1-\epsilon))$ are the audio example ${\mathbf x}_{i}$ and the perturbation vector $\mathbf v$ represented in the $\tanh$ space, respectively, and $\epsilon$ is a small constant that depends on the extreme values of the transformed signal that ensures that ${\mathbf x}_{i}^{\prime}$ and $\mathbf v^{\prime}$  does not assume infinity values. For instance, $\epsilon$=1$e-$7 is a suitable value for the datasets used in Section~\ref{sec:exp_protocol}. The audio example ${\mathbf x}_{i}$ must be transformed to $\tanh$ space and then Eq.~\eqref{eq:trans} can be used to transform the perturbed data to the valid range of $[0, 1]$. Since \(-1 \leq \tanh ({\mathbf x}_{i}^\prime+{\mathbf v^\prime}) \leq 1\) then \(0 \leq {\mathbf w}_{i} \leq 1\) and the solution will be valid according to the box constraint. Refer to Appendix~\ref{app:A_math_step_v^pr} for details. As a result of this transformation, the produced perturbation vector is also in $\tanh$ space, and $\mathbf v$ can be written as: 
\begin{equation}
\label{eq:v_for_v_prime}
\mathbf v= \frac{\tanh(\mathbf v^\prime)+1-\epsilon}{2-2\epsilon}.
\end{equation}

In order to solve the optimization problem defined in Eq.~\eqref{org:prob}, we rewrite variable ${\mathbf v^\prime}$ as follows:
\begin{equation}
\label{eq:v}
{\mathbf v^\prime}=\frac{1}{2}\ln \left ( {\frac{{\mathbf w}_{i}}{1-{\mathbf w}_{i}}}  \right ) -{\mathbf x}_{i}^\prime.
\end{equation}
The details of expressing ${ \mathbf v^\prime}$ as a function of ${\mathbf w}_{i}$ are shown in Appendix~\ref{app:A_math_step}. Therefore, we propose a penalty method that optimizes the following objective function:
\begin{equation}\label{eq:penalty_targeted}
\min_{{\mathbf w}_{i}}
\begin{Bmatrix*}[l]
  L({\mathbf w}_{i},t) =  \text{SPL}\left ( \frac{1}{2}\ln \left ( {\frac{{\mathbf w}_{i}}{1-{\mathbf w}_{i}}}  \right ) -{\mathbf x}_{i}^\prime \right )+c.G({\mathbf w}_{i}, t),
\vspace{5pt}\\
G({\mathbf w}_{i}, t)=\max\{\max\limits_{j\neq t}\left \{ f({\mathbf w}_{i})_{j} \right  \}- f({\mathbf w}_{i})_{t}, -\kappa\}
\end{Bmatrix*}
\end{equation}
\noindent where \(t\) is the target class, \(f({\mathbf w}_{i})_{j}\) is the output of the pre-softmax layer (logit) of a neural network for class \(j\), $c$ is a positive constant known as \textit{"penalty coefficient"} and \(\kappa\) controls the confidence level of sample misclassification. This formulation enables the attacker to control the confidence level of the attack. For untargeted attacks, we modify the objective function of Eq.~\eqref{eq:penalty_targeted} as:
\begin{equation}\label{eq:penalty_untargeted}
\min_{{\mathbf w}_{i}}
\begin{Bmatrix*}[l]
L({\mathbf w}_{i},y_{l}) =  \text{SPL}\left ( \frac{1}{2}\ln \left ( {\frac{{\mathbf w}_{i}}{1-{\mathbf w}_{i}}}  \right ) -{\mathbf x}_{i}^\prime \right )+c.G({\mathbf w}_{i},y_{l}),
\vspace{5pt}\\ 
G({\mathbf w}_{i},y_{l})=\max\{f({\mathbf w}_{i})_{y_l}-\max\limits_{j\neq y_{l}}\left \{ f({\mathbf w}_{i})_{j}  \right \}, -\kappa\}
\end{Bmatrix*}
\end{equation}
\noindent where \(y_{l}\) is the legitimate label for the $i$-th sample of the batch. \(G({\mathbf w}_{i},t)\) is the hinge loss penalty function, which for a targeted attack and $\kappa=0$, must satisfy:
\begin{equation}
\begin{aligned}
\label{eq:hinglossdef}
G({\mathbf w}_{i},t)= 0 \quad \text{if } \quad y_{t}= \argmax_{y}\mathbb{P}(y|{\mathbf w}_{i},\theta),
\\
G({\mathbf w}_{i},t)>0 \quad \text{if }\quad y_{t}\neq \argmax_{y}\mathbb{P}(y|{\mathbf w}_{i}, \theta),
\end{aligned}
\end{equation}
The same properties of the penalty function are also valid for untargeted perturbations. This penalty function is convex and has subgradients therefore, a gradient-based optimization algorithm, such as the Adam algorithm \cite{kingma2014adam} can be used to minimize the finite-sum loss defined in Eqs.~(\ref{eq:penalty_targeted}) and (\ref{eq:penalty_untargeted}). Several other optimization algorithms like AdaGrad \cite{adagrad}, standard gradient descent, gradient descent with Nesterov momentum \cite{nestrov} and RMSProp \cite{Goodfellow-et-al-2016} have also been evaluated but Adam converges in fewer iterations and it produces relatively similar solutions. Algorithm \ref{Univ_pen} presents the pseudo-code of the proposed penalty method.

\textbf{Theorem 1:} Let $\left\{{\mathbf v}^{k}\right\}$, $k=1, ..., \infty$ be the sequence generated by the proposed penalty method in Algorithm 1 for $k$ iterations. Let $ {\bar{\mathbf v}}$ be the limit point of $\left\{\mathbf v^{k}\right\}$. Then any limit point of the sequence is a solution to the original optimization problem defined in Eq.~\eqref{org:prob}\footnote{Theorem 1 applies to the context of convex optimization. The neural network is defined as a functional constraint on the optimization problem defined in Eq.~\eqref{org:prob}. Since neural networks are not convex, a feasible solution to the optimization problem may not be unique and it is not guaranteed to be a global optimum. Moreover, the Theorem 1 is proved based on the assumption defined in Eq.~\eqref{eq:hinglossdef} i.e. $\kappa$=0}.

\textbf{Proof:} According to Eqs.~\eqref{eq:v_for_v_prime} and~\eqref{eq:v}, ${\mathbf v}^{k}$ can be defined as:
\begin{equation}
\label{eq:v_k}
\begin{aligned}
{\mathbf v}^{\prime^k} & =\frac{1}{2}\ln \left ( {\frac{{\mathbf w}^{k}_{i}}{1-{\mathbf w}^{k}_{i}}}  \right ) -{\mathbf x}_{i}^\prime,\\
\mathbf v^{k}&= \frac{\tanh(\mathbf v^{\prime^k})+1-\epsilon}{2-2\epsilon}
\end{aligned}
\end{equation}
Before proving Theorem 1, a useful Lemma is also presented and proved.

\textbf{Lemma 1:} Let ${\mathbf v}^*$ be the optimal value of the original constrained problem defined in Eq.~\eqref{org:prob}. Then $\text{SPL}\left(\mathbf v^{*}\right) \geq L\left({\mathbf w}^{k}_{i}, t\right) \geq \text{SPL}\left(\mathbf v^{k}\right) \forall k$.

\textbf{Proof of Lemma 1:} 
\begin{equation*}
\arraycolsep=1pt
\begin{array}{lll}
\text{SPL}(\mathbf v^{*})&=&{\text{SPL}(\mathbf v^{*})+c.G(\mathbf w^{*}_{i},t)} \quad (\because {G(\mathbf w^{*}_{i},t) = 0)} \\[1pt]
&\geq&\text{SPL}(\mathbf v^{k})+c.G(\mathbf w^{k}_{i},t) \quad (\because c>0, G(\mathbf w^{k}_{i},t) \geq 0,\\[1pt]
& & \hfill \mathbf w^{k}_{i} \,\text{minimizes} \ L({\mathbf w}^{k}_{i},t))\\[1pt]
&\geq&\text{SPL}({\mathbf v}^{k}) \\[1pt]
\therefore \text{SPL}({\mathbf v}^{*})&\geq& L({\mathbf w}^{k}_{i}, t) \geq \text{SPL}({\mathbf v}^{k}) \, \forall k.
\end{array}
\end{equation*}

\textbf{Proof of Theorem 1}. $\text{SPL}$ is a monotonically increasing function and continuous. Also, $G$ is a hinge function, which is continuous. 
$L$ is the summation of two continuous functions. Therefore, it is also a continuous function. The limit point of $\left\{{\mathbf v}^{k}\right\}$ is defined as: $\bar{\mathbf v}= \lim_{k\rightarrow \infty} {\mathbf v}^{k}$ and since $\text{SPL}$ is a continuous function, $\text{SPL}(\bar{\mathbf v})= \lim_{k\rightarrow \infty} \text{SPL}({\mathbf v}^{k})$. We can conclude that:
\begin{equation*}
\begin{array}{llll}
L^{*} &=& \lim_{k\rightarrow \infty} L({\mathbf w}^{k}_{i}, t) \leq \text{SPL}({\mathbf v}^{*})  \quad (\because \text{Lemma 1}) \\[1pt]
L^{*} &=& \lim_{k\rightarrow \infty} \text{SPL}({\mathbf v}^{k})+ \lim_{k\rightarrow \infty} c.G(\mathbf w^{k}_{i},t) \leq \text{SPL}({\mathbf v}^{*})\\[1pt]
L^{*} &=& \text{SPL}(\bar{\mathbf v})+\lim_{k\rightarrow \infty} c.G(\mathbf w^{k}_{i},t) \leq \text{SPL}({\mathbf v}^{*}).
\end{array}
\end{equation*}
If ${\mathbf v}^k$ is a feasible point for the constrained optimization problem defined in Eq.~\eqref{org:prob}, then, from the definition of function $G(.)$, one can conclude
that $\lim_{k\rightarrow \infty} c.G(\mathbf w^{k}_{i},t)= 0$. Then:
\begin{equation*}
L^{*} = \text{SPL}(\bar{\mathbf v}) \leq \text{SPL}({\mathbf v}^{*})\\[1pt]
\end{equation*}
\begin{equation*}
\boxed{\therefore \, \bar{\mathbf v} \, \text{is a solution of the problem defined in Eq.~\eqref{org:prob}}} 
\end{equation*}

\begin{algorithm}[ht]
\KwIn{Data points \(X=\left\{{\mathbf x}_{1}, \dots, {\mathbf x}_{m}\right\}\) with corresponding legitimate labels \(Y\), desired fooling rate on perturbed samples \(\delta\), and target class \(t\) (for targeted attacks)}
\KwOut{Universal perturbation signal \(\mathbf v^\prime\)}
\nl \bf initialize \(\mathbf v^\prime \leftarrow 0\), \\
\nl \bf \While {$\operatorname{ASR}\left(X,\mathbf v^\prime\right) \leq  1 - \delta$}{
        \nl \bf {\normalfont Sample a mini-batch of size $S$ from $(X,Y)$}\\
        \nl \bf ${\mathbf g} \leftarrow 0$\\
       \nl \bf \For{ $i \leftarrow 1 \; to \; S$}    
        { 
            \nl \bf {\normalfont Transform the audio signal to $\tanh$ space}:\\
        	        \({\mathbf x}_{i}^\prime = \arctanh\left ((2{\mathbf x}_{i}-1)*(1-\epsilon)\right)\),\\
        	        
        	 \nl \bf {\normalfont Compute the transformation of the perturbed signal for each sample $i$ from mini-batch}:\\
        	        \({\mathbf w}_{i}=\frac{1}{2}(\tanh ({\mathbf x}_{i}^\prime+\mathbf v^\prime)+1)\),\\
             {\normalfont Compute the gradient of the objective function, i.e., Eq.~\eqref{eq:penalty_targeted} or Eq.~\eqref{eq:penalty_untargeted}, w.r.t. \({\mathbf w}_{i}\)}: \\
    \nl \bf  \eIf{targeted attack:}{
           \nl \bf \({\mathbf g}\leftarrow {\mathbf g}+ \frac{\partial L({\mathbf w}_{i},t)}{\partial {\mathbf w}_{i}}\)
                   }{
           \nl \bf \({\mathbf g}\leftarrow {\mathbf g}+ \frac{\partial L({\mathbf w}_{i},y_{i})}{\partial {\mathbf w}_{i}}\)
          }
        }
        \nl \bf {\normalfont Compute update \(\Delta \mathbf v^\prime\) using \(\mathbf g\) according to Adam update rule \cite{kingma2014adam}} \\
        \nl \bf {\normalfont apply update}:  \( \mathbf v^\prime \leftarrow \mathbf v^\prime+ \Delta \mathbf v^\prime  \)
        }
        \Return $\mathbf v$
    \caption{{\bf Penalty method for universal adversarial audio perturbations.} \label{Univ_pen}}
\end{algorithm}

\section{Experimental Protocol and Results}
\label{sec:exp_protocol}
The proposed UAP is evaluated on two audio tasks: environmental sound classification and speech command recognition. For environmental sound classification, we have used the full audio recordings of UrbanSound8k dataset~\cite{Salamon:2014:DTU:2647868.2655045} downsampled to 16 kHz for training and evaluating the models as well as for generating adversarial perturbations. This dataset consists of 7.3 hours of audio recordings split into 8,732 audio clips of up to slightly more than three seconds. Therefore, each audio sample is represented by a 50,999-dimensional array. The audio clips were categorized into 10 classes. The dataset was split into training (80\%), validation (10\%) and test (10\%) set. For generating perturbations, 1,000 samples of the training set were randomly selected, and for penalty-based method a mini-batch size of 100 samples is used. The perturbations were evaluated on the whole test set (874 samples).

For speech command recognition, we have used the Speech Command dataset, which consists of 61.83 hours of audio sampled at 16 kHz \cite{warden2018speech}, and categorized into 32 classes. The training set  consists of 17.8 hours of speech command recordings split into 64,271 audio clips of one second. The test set consists of 158,537 audio samples corresponding to 44.03 hours of speech commands. This dataset was used as the benchmark dataset for TensorFlow Speech Recognition (TFSR) challenge in 2017\footnote{https://www.kaggle.com/c/tensorflow-speech-recognition-challenge}. This challenge was based on the principles of open science so, data, source code, and evaluation methods of over 1,300 participant teams are publicly available for commercial and non-commercial usage. So, as it is shown in this study, it is quite straightforward for the adversary to attack the model.
\begin{figure*}[htpb!]%
\centering
\subfigure[]{%
\includegraphics[width=.21\textwidth]{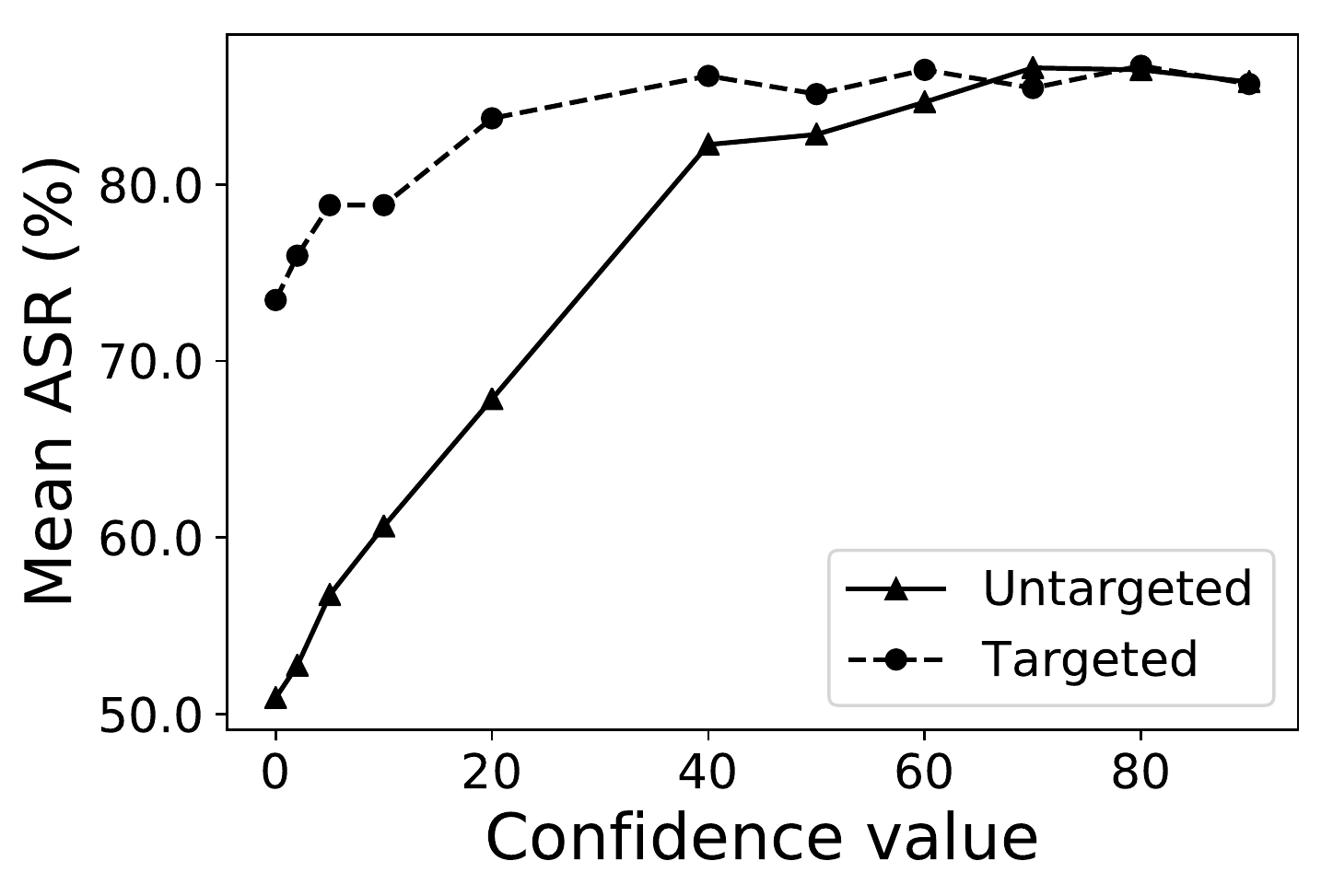}}%
\subfigure[]{%
\includegraphics[width=.21\textwidth]{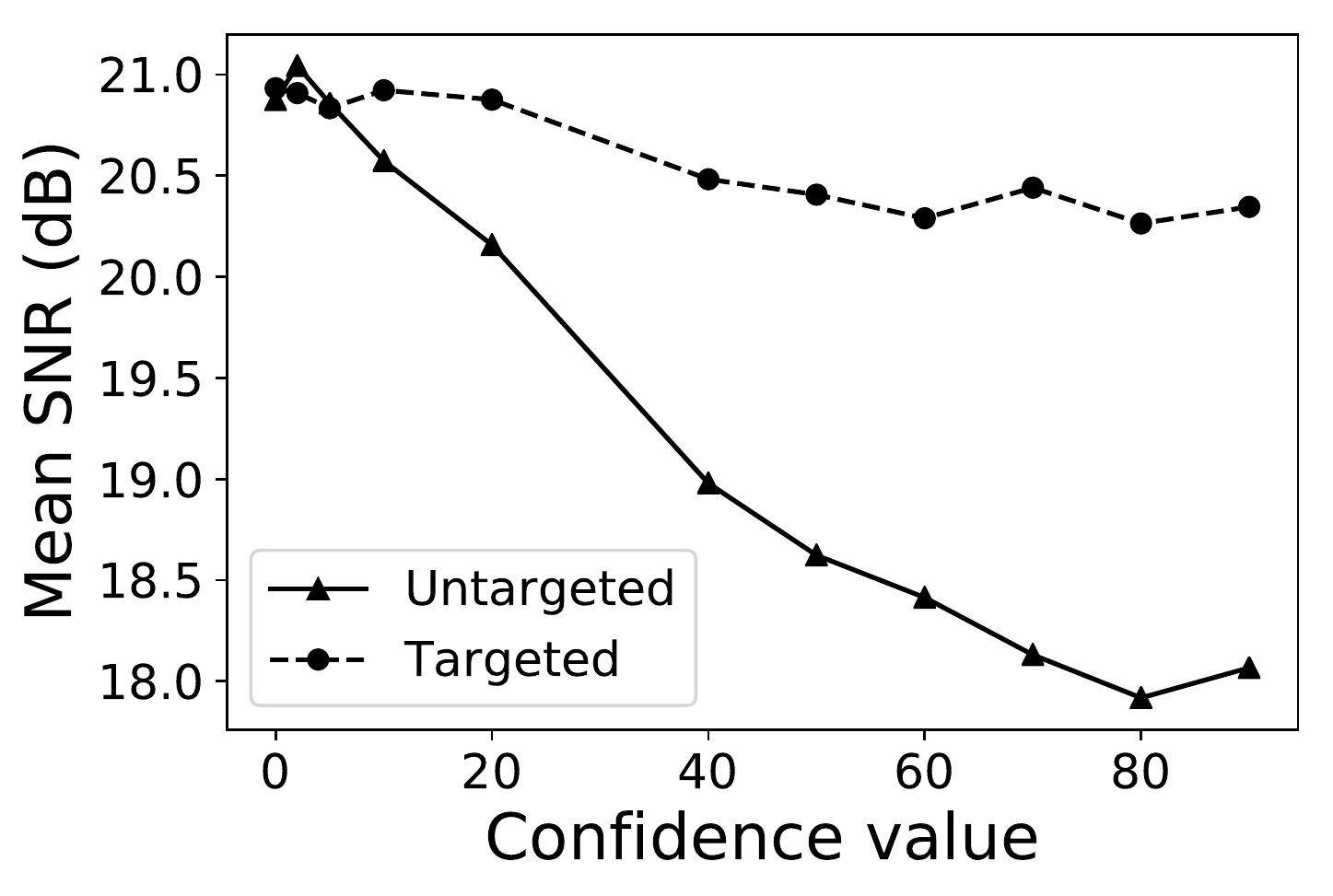}}%
\subfigure[]{%
\includegraphics[width=.21\textwidth]{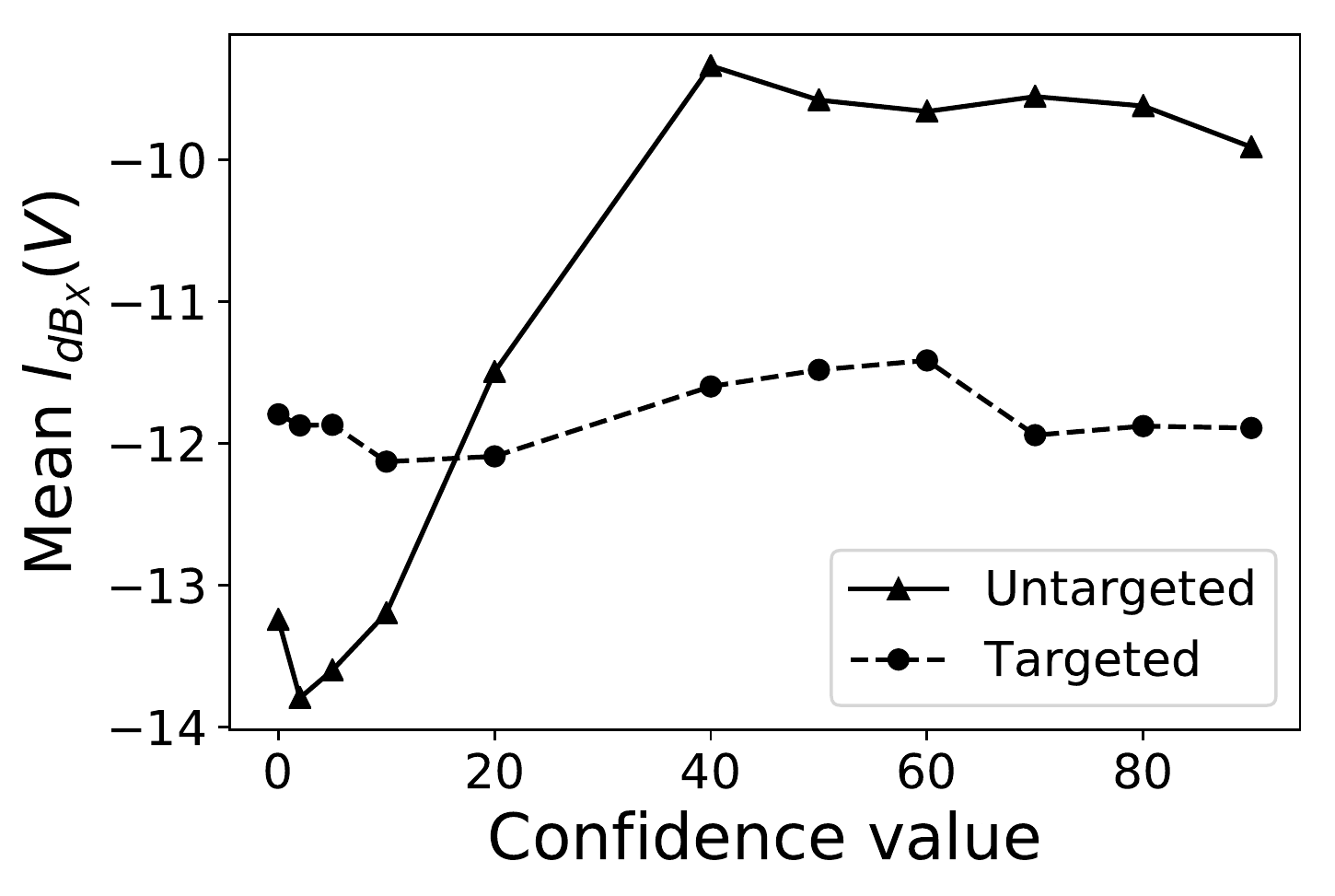}}\\%
\subfigure[]{%
\includegraphics[width=.21\textwidth]{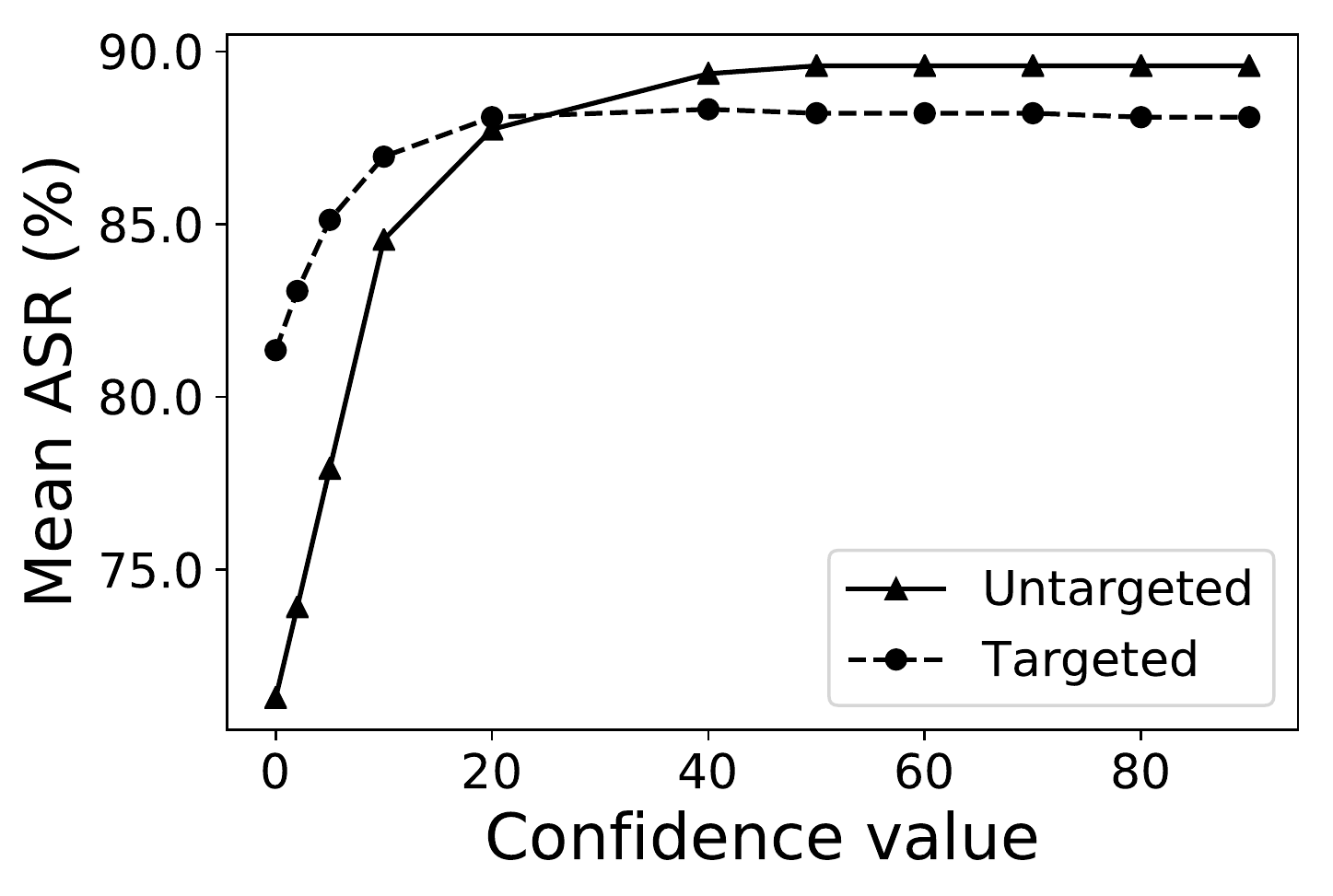}}%
\subfigure[]{%
\includegraphics[width=.20\textwidth]{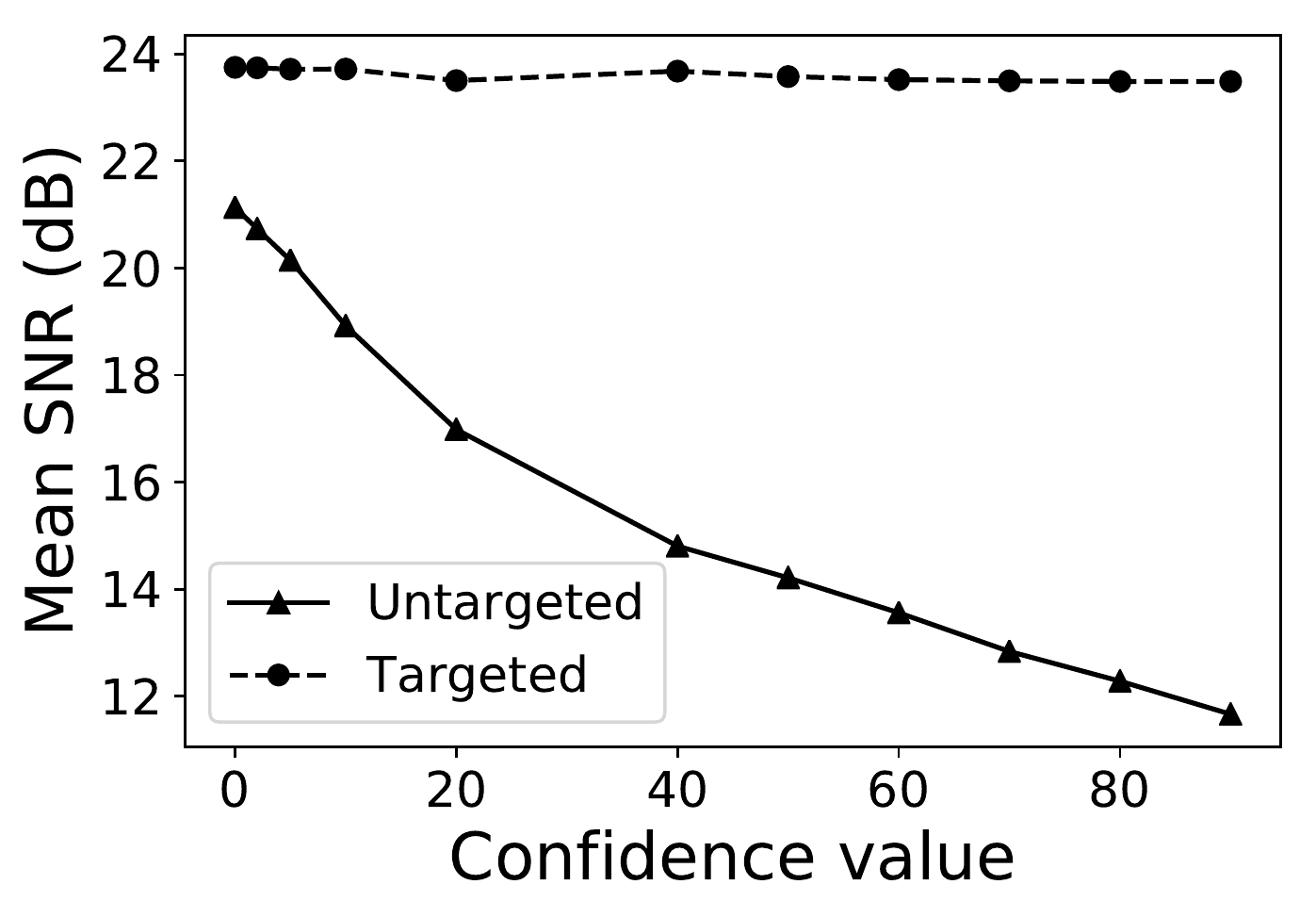}}%
\subfigure[]{%
\includegraphics[width=.21\textwidth]{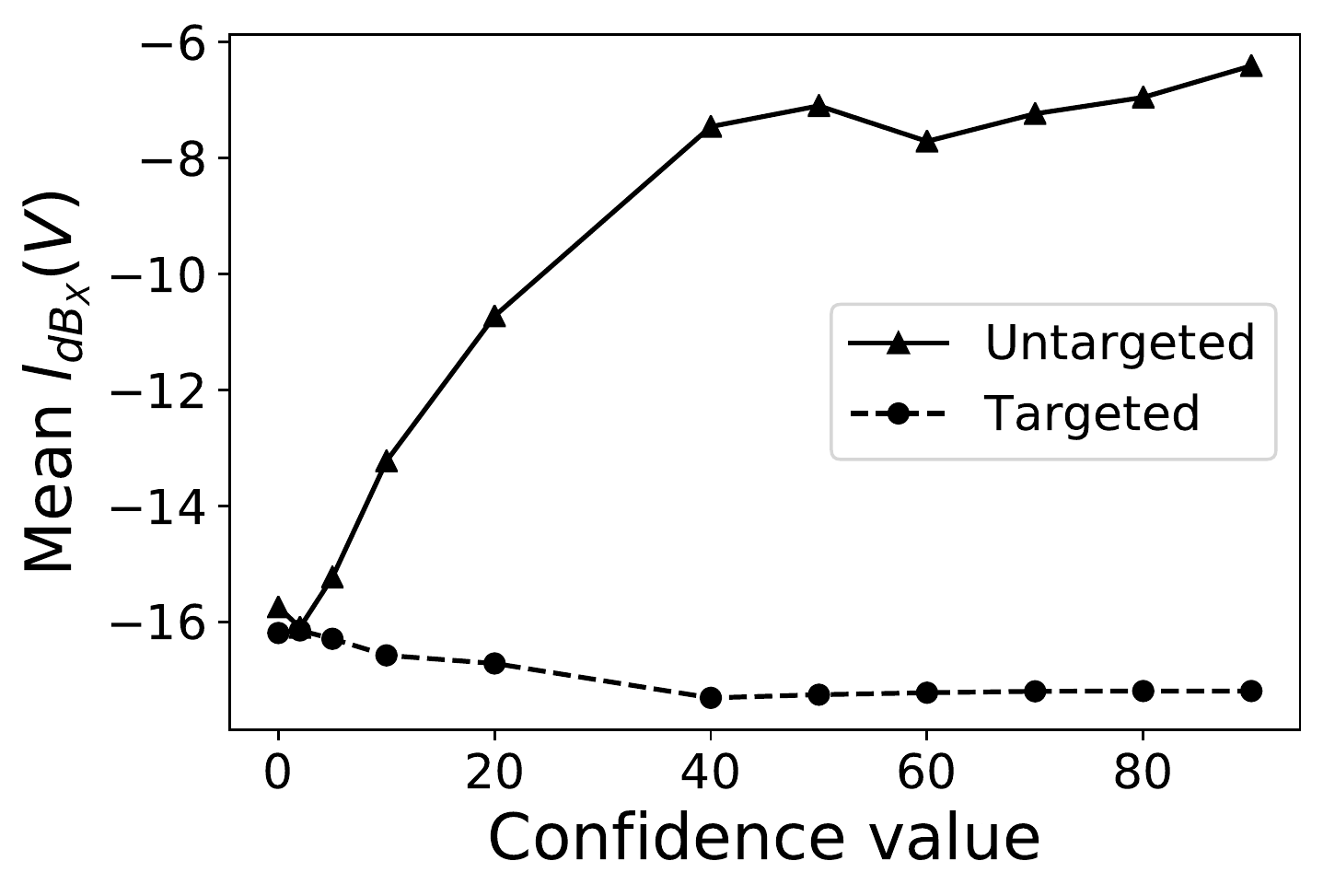}}
\caption{{Effect of different confidence values on the mean values of ASR, SNR and $l_{\text{dB}_{\mathbf x}}(\mathbf v)$ for targeted and untargeted attacks. Top row (a, b, c): ENVnet-V2 \cite{tokozume2017learning} used as target model. Bottom row (d, e, f): 1D CNN Gamma \cite{abdoli2019end} used as target model.}}
\label{fig:conf_val}
\end{figure*}

Signal-to-noise Ratio (SNR) is used as a metric to measure the level of noise with respect to the original signal. This metric, which is also measured in dB, is used for measuring the level of the perturbation of the signal after adding the universal perturbation. This measure is also used in previous works for evaluating the quality of the generated adversarial audio attacks \cite{kereliuk2015deep,du2019sirenattack}, and it is defined as:
\begin{equation}
\text{SNR}({\mathbf x},{\mathbf v})=20\log_{10}{ \frac{P({\mathbf x})}{P(\mathbf v)}},
\end{equation}
\noindent where \(P(.)\) is the power of the signal defined in Eq.~\eqref{eq:power}. A high SNR indicates that a low level of noise is added to the audio sample by the universal adversarial perturbation. Additionally, the relative loudness of perturbation with respect to audio sample, measured in $dB$, is also applied:
\begin{equation}
\label{eq:MeandB}
l_{\text{dB}_{\mathbf x}}(\mathbf v)=l_{\text{dB}}(\mathbf v)-l_{\text{dB}}(\mathbf x)
\end{equation}
where $l_{\textit dB}(\mathbf x) = \max_{n}(20\log_{10}(x_{n}))$ and  $x_{n}$ denotes the $n$-th component of the array ${\mathbf x}$. This measure is similar to $\ell_{\infty}$ norm in image domain. Smaller values indicate quieter distortions. This measure is also applied in recent studies for quality assessment of adversarial attacks against audio processing models \cite{Neekhara_2019, carlini2018audio, yang2018characterizing}.

We have chosen a family of diverse end-to-end architectures as our target models. This selection is based on choosing architectures which might learn representations directly from the audio signal. We briefly describe the architecture of each model as follows. 1D CNN Rand, 1D CNN Gamma, ENVnet-V2, SincNet and  SincNet+VGG19 are just used for environmental sound classification while SpchCMD model is used for speech command recognition. A detailed description of the architectures can be found in supplementary material.

\textbf{1D CNN Rand~\cite{abdoli2019end}:} This model consists of five one-dimensional convolutional layers (CLs). The output of CLs is used as input to two fully connected (FC) layers followed by an output layer with softmax activation function. The weights of all of the layers are initialized randomly. This model was proposed by Abdoli \etal\cite{abdoli2019end} for environmental sound classification.

\textbf{1D CNN Gamma~\cite{abdoli2019end}:} This model is similar to 1D CNN Rand except that it employs a Gammatone filter-bank in its first layer. Furthermore, this layer is kept frozen during the training process. Gammatone filters are used to decompose the input signal to appropriate frequency bands. 

\textbf{ENVnet-V2~\cite{tokozume2017learning}:} The architecture
for sound recognition was slightly modified to make it compatible with the input size of the downsampled audio samples of UrbanSound8k dataset. This architecture uses the raw audio signal as input, and it extracts short-time frequency features by using two one-dimensional CLs followed by a pooling layer (PL). It then swaps axes and convolves in time and in frequency domain the features using five two-dimensional CLs. Two FC layers and an output layer with softmax activation function complete the network.

\textbf{SincNet~\cite{ravanelli2018speaker}:} The end-to-end architecture
for sound processing extracts meaningful features from the audio signal at its first layer. In this model, several sinc functions are used as band-pass filters and only low and high cutoff frequencies are learned from audio. After that, two one-dimensional CLs are applied. Two FC layers followed by an output layer with softmax activation are used for classification.

\textbf{SincNet+VGG19~\cite{ravanelli2018speaker}:} This model uses sinc filters to extract features from the raw audio signal as in SincNet~\cite{ravanelli2018speaker}. After an one-dimensional maxpooling layer, the output is stacked along time axis to form a 2D representation. This time-frequency representation is used as the input to a VGG19 network~\cite{VGG19} followed by a FC layer and an output layer with softmax activation for classification. This time-frequency representation resembles a spectrogram representation of the audio signal.

\textbf{SpchCMD}: This architecture was proposed by the winner of the TFSR challenge. The model is based on a CNN, which uses one-dimensional CLs and several depth-wise CLs for extracting useful information from several chunks of raw audio. The model also uses an attention layer and a global average PL. According to the challenge rules, the model must handle silence signal and also unknown samples from other proposed classes in Speech Command dataset \cite{warden2018speech}. Therefore, the softmax layer has 32 outputs.

For the iterative method, several parameters must be chosen. In order to find the minimal perturbation $\Delta {\mathbf v}_{i}$, we used the DDN $\ell_2$ attack \cite{rony2018decoupling}. This attack is designed to efficiently find small perturbations that fool the model. The difference with DeepFool \cite{moosavi2016deepfool} is that it can be used for both untargeted and targeted attacks, extending the iterative method to the targeted scenario. DDN was used with a budget of $50$ steps and an initial norm of $0.2$. Results are reported for $p$=$\infty$ and we set $\xi$ to 0.2 and 0.12 for untargeted and targeted attack scenarios, respectively. These values were chosen to craft perturbations in which the norm is much lower than the norm of the audio samples in the dataset.

For evaluating the penalty method we set the penalty coefficient $c$ to 0.2 and 0.15 for untargeted and targeted attack scenarios, respectively. The confidence value $\kappa$ is set to 40 and 10 for crafting untargeted and targeted perturbations, respectively. For both methods, based on our initial experiments, we found that these values are appropriate to produce fine quality perturbed samples within a reasonable number of iterations. Fig.~\ref{fig:conf_val} shows the effect of different confidence values on mean ASR, mean SNR and mean $l_{\textit dB_{\mathbf x}}(\mathbf v)$ for targeted and untargeted attacks on the ENVnet-V2 \cite{tokozume2017learning} and 1D CNN Gamma \cite{abdoli2019end} models for the test set. For this experiment, 1,000 and 500 training samples for untargeted and targeted attack scenarios are used, respectively. For targeted attacks, the target class is \textit{"Gun shot"}. Fig.~\ref{fig:conf_val} shows that the ASR increases as confidence value increases. However, the SNR also decreases in the same way. For untargeted attacks, it has a more detrimental effect on the mean $l_{\textit dB_{\mathbf x}}(\mathbf v)$ than targeted scenario. For both iterative and penalty methods, we set the desired fooling rate on perturbed training samples to $\delta$=0.1. Both algorithms terminate execution whether they achieve the desired fooling ratio, or they reach 100 iterations. Both algorithms have been trained and tested using a TITAN Xp GPU.
\begin{center}
\begin{table*}[htpb!]
\caption{Mean values of ASR, SNR and $l_{\text{dB}_{\mathbf x}}(\mathbf v)$ on training and test sets for untargeted and targeted perturbations. Higher ASRs are in boldface.}
\centering
\begin{tabular}{clcccm{5em}cccm{5em}}
\toprule
                           &               & \multicolumn{4}{c}{Targeted Attack} & \multicolumn{4}{c}{Untargeted Attack} \\ 
            \cmidrule(l){3-6}\cmidrule(l){7-10} 
                           &               & Training Set & \multicolumn{3}{c}{Test Set} & Training Set & \multicolumn{3}{c}{Test Set} \\ 
           \cmidrule(l){3-3}\cmidrule(l){4-6}\cmidrule(l){7-7}\cmidrule(l){8-10}
Method                     & Model         & ASR & ASR & SNR (dB) & $l_{\text{dB}_{\mathbf x}}(\mathbf v)$ & ASR & ASR & SNR (dB) & $l_{\text{dB}_{\mathbf x}}(\mathbf v)$   \\ 
\midrule
\multirow{5}{*}{Iterative} & 1D CNN Rand   & 0.926 & 0.672 & 25.321 & -18.416  & 0.911 & 0.412 & 25.244  & -14.258   \\
                           & 1D CNN Gamma  & 0.945 & 0.795 & 23.587 & -18.416  & 0.904 & 0.737 & 22.922  & -13.979   \\
                           & ENVnet-V2     & 0.916 & 0.767 & 22.734 & -18.416  & 0.910 & 0.669 & 24.960  & -13.979   \\
                           & SincNet       & 1.000 & 0.899 & 28.668 & -21.044  & 0.915 & 0.886 & 24.025  & -18.959  \\
                           & SincNet+VGG19 & 0.985 & 0.872 & 26.203 & -18.416  & 0.924 & 0.838 & 26.362  & -14.007  \\ \midrule
\multirow{5}{*}{Penalty}   & 1D CNN Rand   & 0.917 & \textbf{0.854} & 23.468 & -16.762 & 0.900 & \textbf{0.876} & 20.350 & -13.371   \\
                           & 1D CNN Gamma  & 0.913 & \textbf{0.888} & 22.835 & -17.375 & 0.901 & \textbf{0.858} & 20.551 & -18.133  \\
                           & ENVnet-V2     & 0.922 & \textbf{0.877} & 21.832 & -14.198 & 0.900 & \textbf{0.831} & 18.727 & -10.490 \\
                           & SincNet       & 0.962 & \textbf{0.971} & 30.411 & -31.916 & 0.900 & \textbf{0.919} & 29.972 & -27.494   \\
                           & SincNet+VGG19 & 0.916 & \textbf{0.898} & 26.736 & -21.059 & 0.902 & \textbf{0.865} & 23.555 & -17.759   \\
\bottomrule                           
\end{tabular}
\label{tab:fooling_ratio_all}
\end{table*}
\end{center}
\begin{center}
\begin{table*}[htpb!]
\caption{Mean values of ASR, SNR and $l_{\text{dB}_{\mathbf x}}(\mathbf v)$ on training and test sets for untargeted and targeted perturbations for targeting speech recognition model (SpchCMD). Higher ASRs are in boldface.}
\centering
\begin{tabular}{ccccm{5em}cccm{5em}m{5em}}
\toprule
                                  & \multicolumn{4}{c}{Targeted Attack} & \multicolumn{5}{c}{Untargeted Attack} \\ \cmidrule(l){2-5} \cmidrule(l){6-10}
                                  & Training Set & \multicolumn{3}{c}{Test Set} & Training Set & \multicolumn{4}{c}{Test Set} \\ \cmidrule(l){2-2}\cmidrule(l){3-5}\cmidrule(l){6-6}\cmidrule(l){7-10}
Method                            & ASR & ASR & SNR (dB) & $l_{\text{dB}_{\mathbf x}}(\mathbf v)$ &  ASR & ASR & SNR (dB) & $l_{\text{dB}_{\mathbf x}}(\mathbf v)$ &  Model Acc.  \\ 
\midrule
\multirow{1}{*}{Iterative}       & 0.902 & \textbf{0.855} & 27.437 & -18.924 & 0.901 &  0.834 & 28.524 & -18.418  & 0.192    \\
\midrule
\multirow{1}{*}{Penalty}         & 0.903  & 0.850 & 26.728 & -24.575 & 0.910 &  \textbf{0.875} & 26.716 & -21.462   & 0.191  \\
\bottomrule                           
\end{tabular}
\label{tabl:fooling_ratio_all_speech}
\end{table*}
\end{center}

\subsection{Results}
\label{sec:results}
Table~\ref{tab:fooling_ratio_all} shows the results of the iterative and penalty methods against the five environmental sound classification models considered in this study. We evaluate both targeted and untargeted attacks in terms of mean ASR on the training and test sets, as well as mean SNR and mean $l_{\text{dB}_{\mathbf x}}(\mathbf v)$ of the perturbed samples of the test set. For crafting the perturbation vector 1,000 randomly chosen samples of the training set is used. For both untargeted and targeted attacks, the penalty method produces the highest ASR for all target models. Both methods produce relatively similar mean SNR on the test set. For targeted scenario, the penalty method produces better results in terms of mean $l_{\text{dB}_{\mathbf x}}(\mathbf v)$ for attacking SincNet and SincNet+VGG19 models. The iterative method, however, works slightly better for attacking ENVnet-V2 model. For the rest of the models, the difference is negligible. The projection operation on $\ell_{\infty}$ ball used in the iterative method is like clipping the perturbation vector. Since the maximum amplitude of the perturbation vector surpasses the limit ($\xi$= 0.12) while producing the perturbation, the projection operation causes the method to achieve the same mean $l_{\text{dB}_{\mathbf x}}(\mathbf v)$ for all methods in this attacking scenario. For the untargeted scenario, the penalty method also produces quieter perturbations in terms of mean $l_{\text{dB}_{\mathbf x}}(\mathbf v)$ for attacking 1D CNN Gamma, SincNet and SincNet+VGG19 models. The crafted universal perturbations produced from the training set by the penalty method generalize better than those produced by the iterative method for both attacking scenarios. We also observe a relatively low difference in ASR between the training and test sets. The detailed results of the targeted attack scenario for each model is reported in supplementary material. For a better assessment of the perturbations produced by both proposed methods, several randomly chosen examples of perturbed audio samples are presented in supplementary material.

Table~\ref{tabl:fooling_ratio_all_speech} shows the results achieved by iterative and penalty methods for targeting the SpchCMD model. In order to find the universal perturbation vector, 3,000 samples from the training set were randomly selected. All samples of the test set were used to evaluate the effectiveness of the perturbation vector. For targeting this model, the perturbation generated by the penalty method is also projected around the $\ell_{2}$ ball according to Eq.~\eqref{eq:proj} with radius $\xi$=6. Based on the preliminary experiments for this task, this projection produces slightly better results in terms of mean SNR and mean $l_{\text{dB}_{\mathbf x}}(\mathbf v)$. For the targeted attack scenario, both methods produce similar results in terms of ASR and SNR however, penalty method generates higher-quality perturbation vectors in terms of mean $l_{\text{dB}_{\mathbf x}}(\mathbf v)$. For untargeted attacks, the penalty method produces better results in terms of mean ASR and mean $l_{\text{dB}_{\mathbf x}}(\mathbf v)$ and both methods produce similar perturbations in terms of SNR. Roughly speaking, SNRs higher than 20 are considered as acceptable ones. As mentioned by Yang \etal\cite{yang2018characterizing}, perturbed audio samples which have $l_{\text{dB}_{\mathbf x}}(\mathbf v)$ ranging from -15 dB to -45 dB are tolerable to human ears. From Tables~\ref{tab:fooling_ratio_all} and~\ref{tabl:fooling_ratio_all_speech}, the mean SNR and $l_{\text{dB}_{\mathbf x}}(\mathbf v)$ of the perturbed samples are mostly in this range. Neekhara \etal\cite{Neekhara_2019} also reported relatively similar results (between -29.82 dB and -41.86 dB) for attacking a speech-to-text model in an untargeted scenario. The model predictions after adding the perturbations were submitted to the evaluation system of the TFSR challenge to evaluate the impact of the perturbations on the model accuracy. Table~\ref{tabl:fooling_ratio_all_speech} shows that the accuracy of the winner speech recognition model (Model Acc.) has dropped to near of random guessing for the perturbations produced by both proposed algorithms. The success rates reported in Tables~\ref{tab:fooling_ratio_all} and~\ref{tabl:fooling_ratio_all_speech} are also statistically significant. Refer to supplementary material for details.
\begin{figure*}[ht]%
\centering
\subfigure[][]{%
\label{fig:ex3-a}%
\includegraphics[width=.20\textwidth]{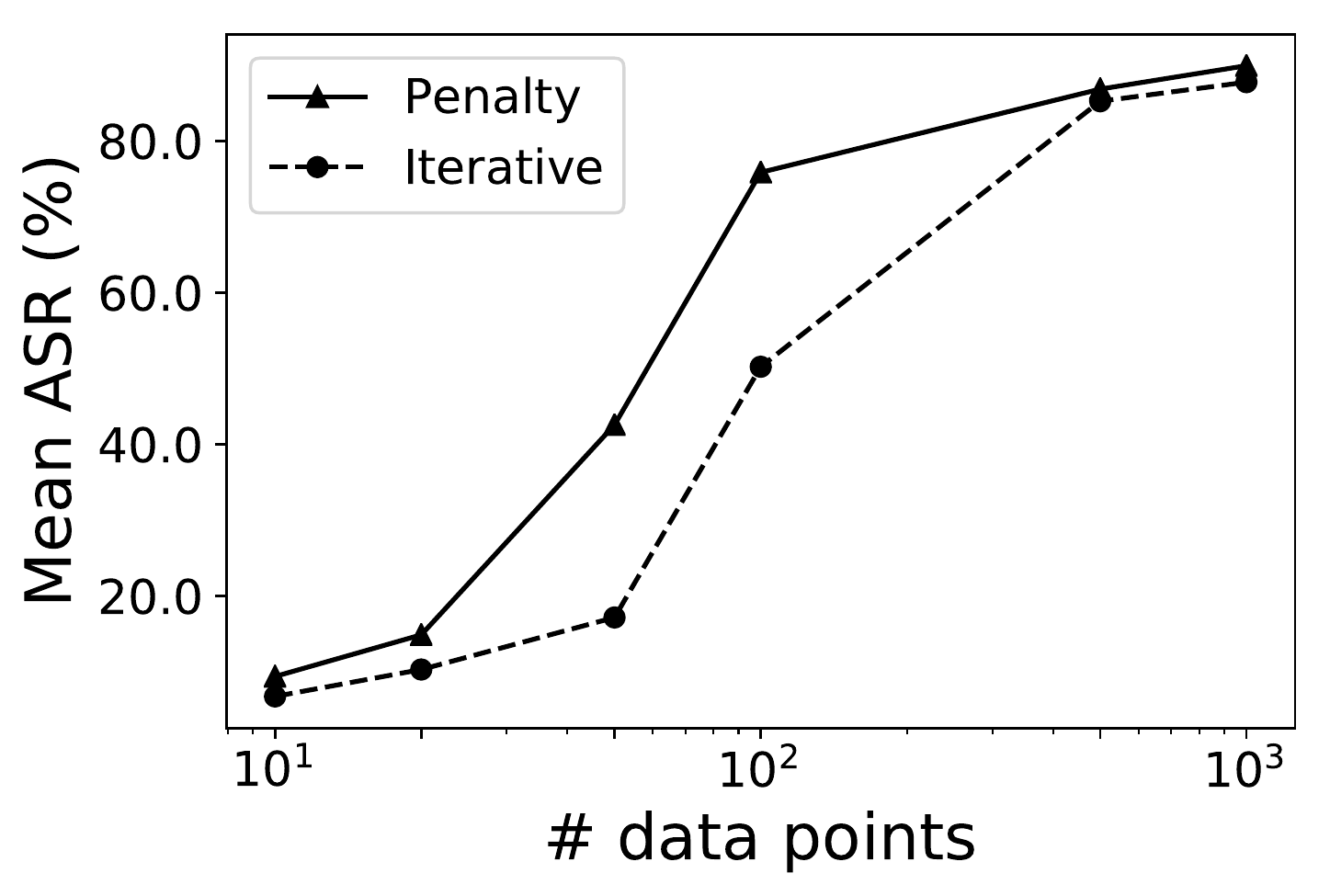}}%
\hspace{1pt}%
\subfigure[][]{%
\label{fig:ex3-b}%
\includegraphics[width=.20\textwidth]{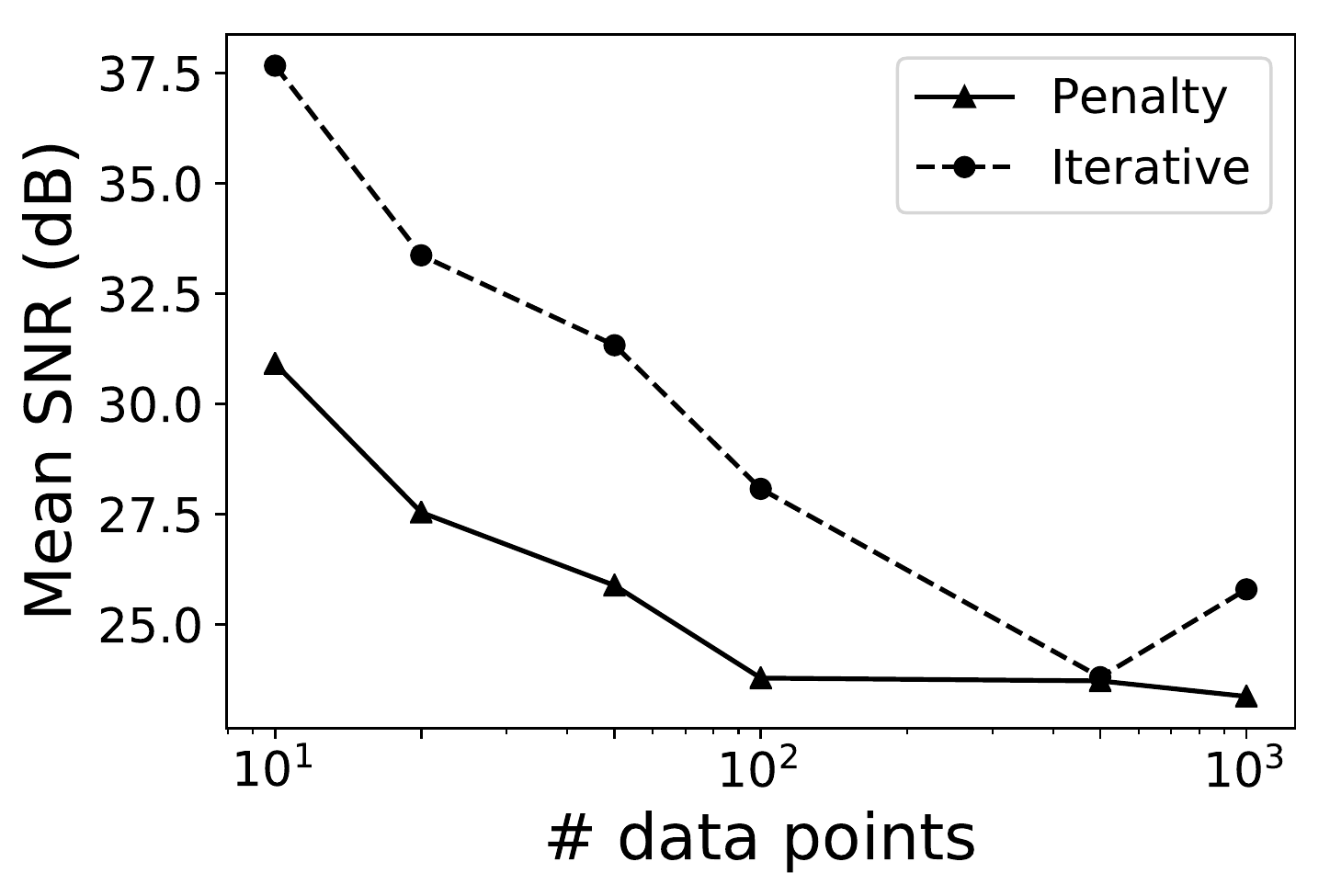}} 
\subfigure[][]{%
\label{fig:ex3-c}%
\includegraphics[width=.20\textwidth]{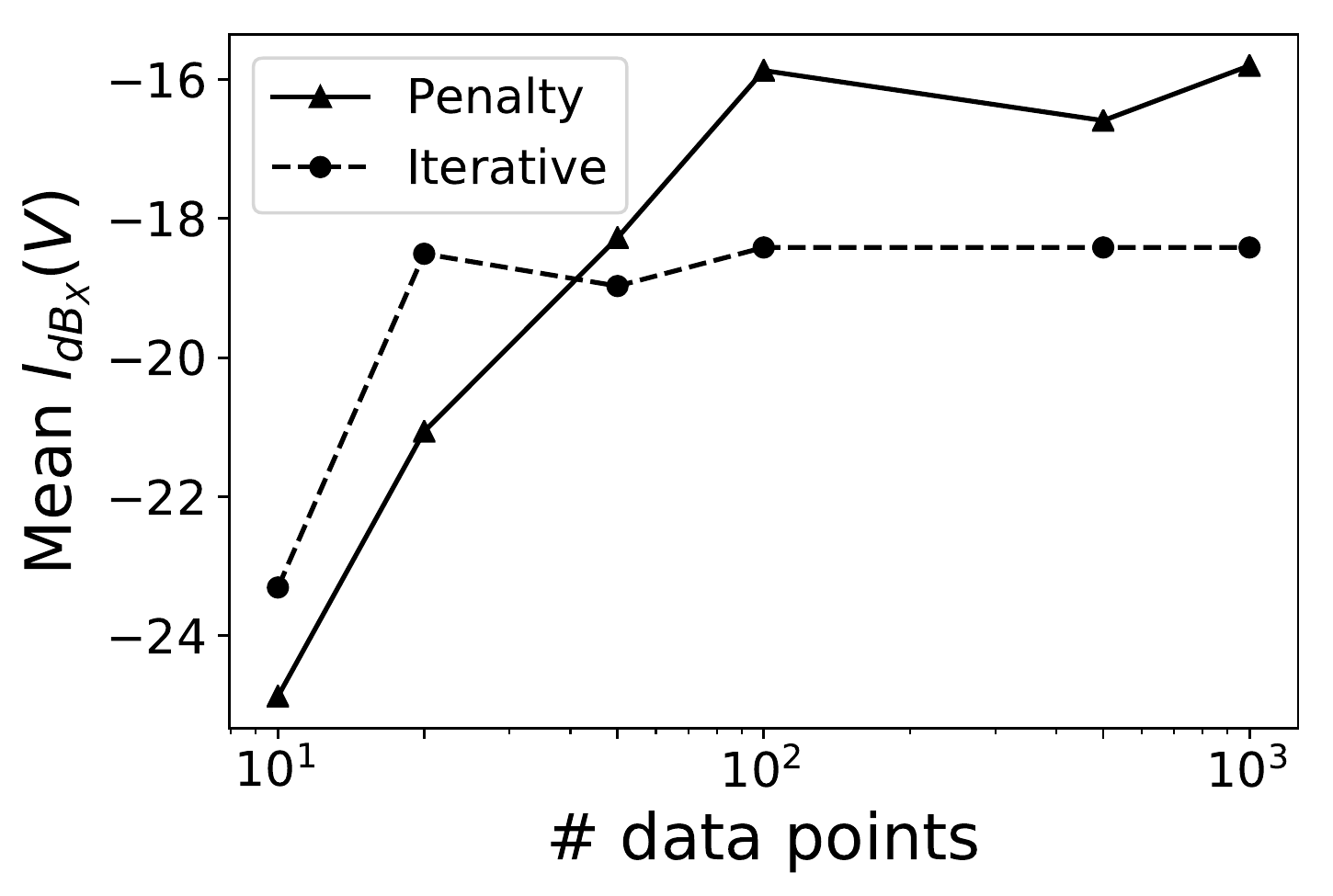}}\\
\subfigure[][]{%
\label{fig:ex3-d}%
\includegraphics[width=.20\textwidth]{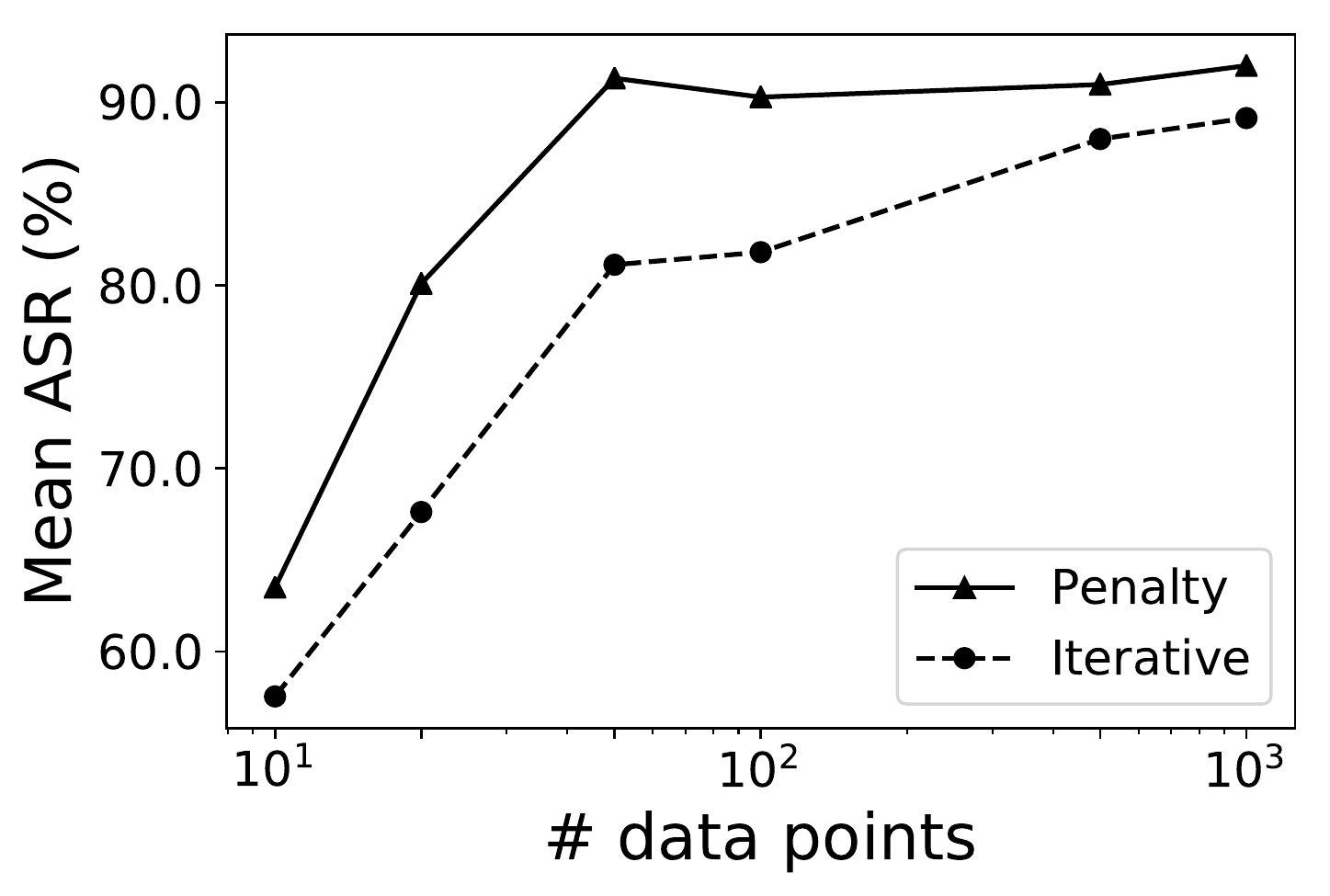}}%
\hspace{1pt}%
\subfigure[][]{%
\label{fig:ex3-e}%
\includegraphics[width=.19\textwidth]{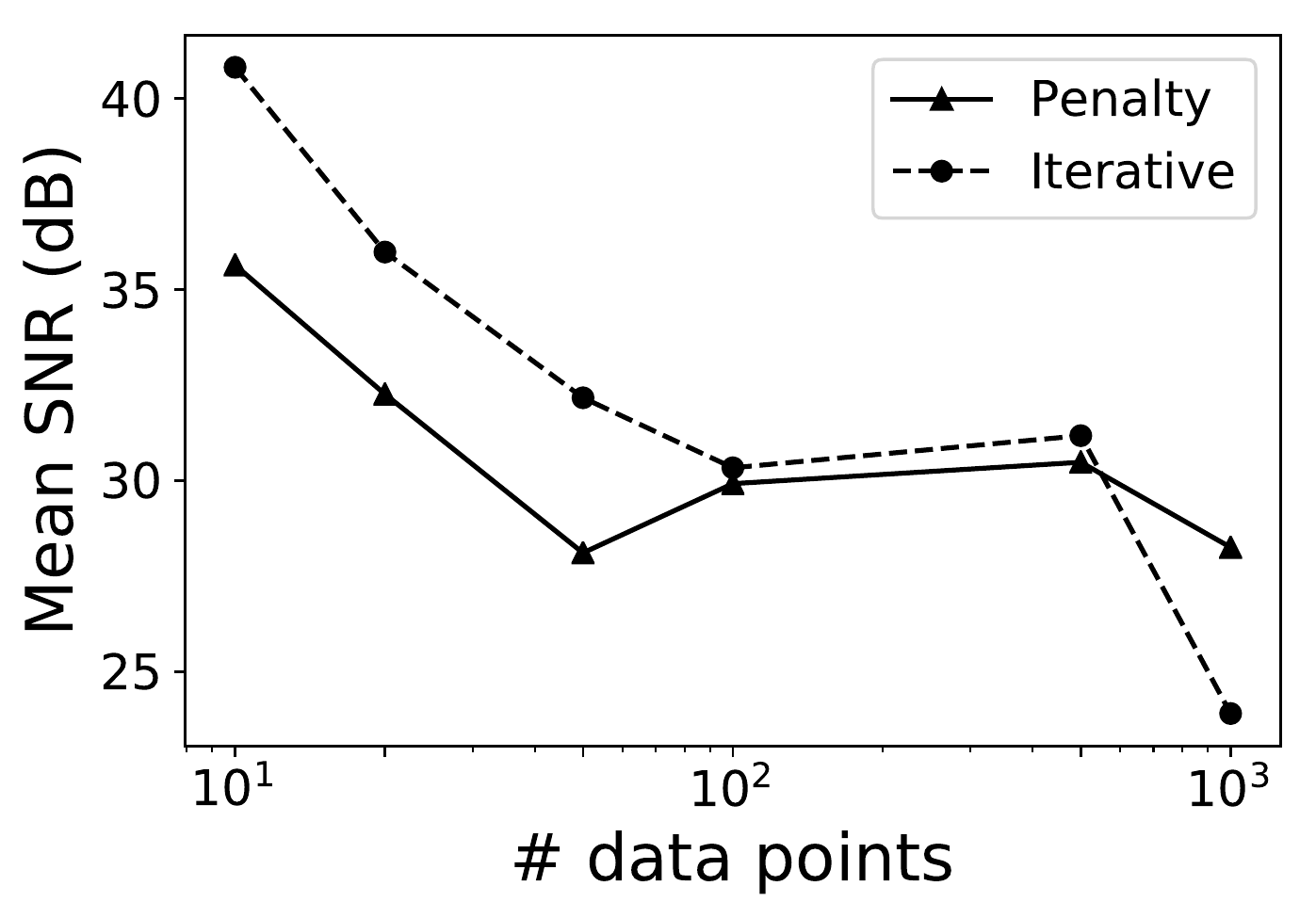}}
\subfigure[][]{%
\label{fig:ex3-f}%
\includegraphics[width=.20\textwidth]{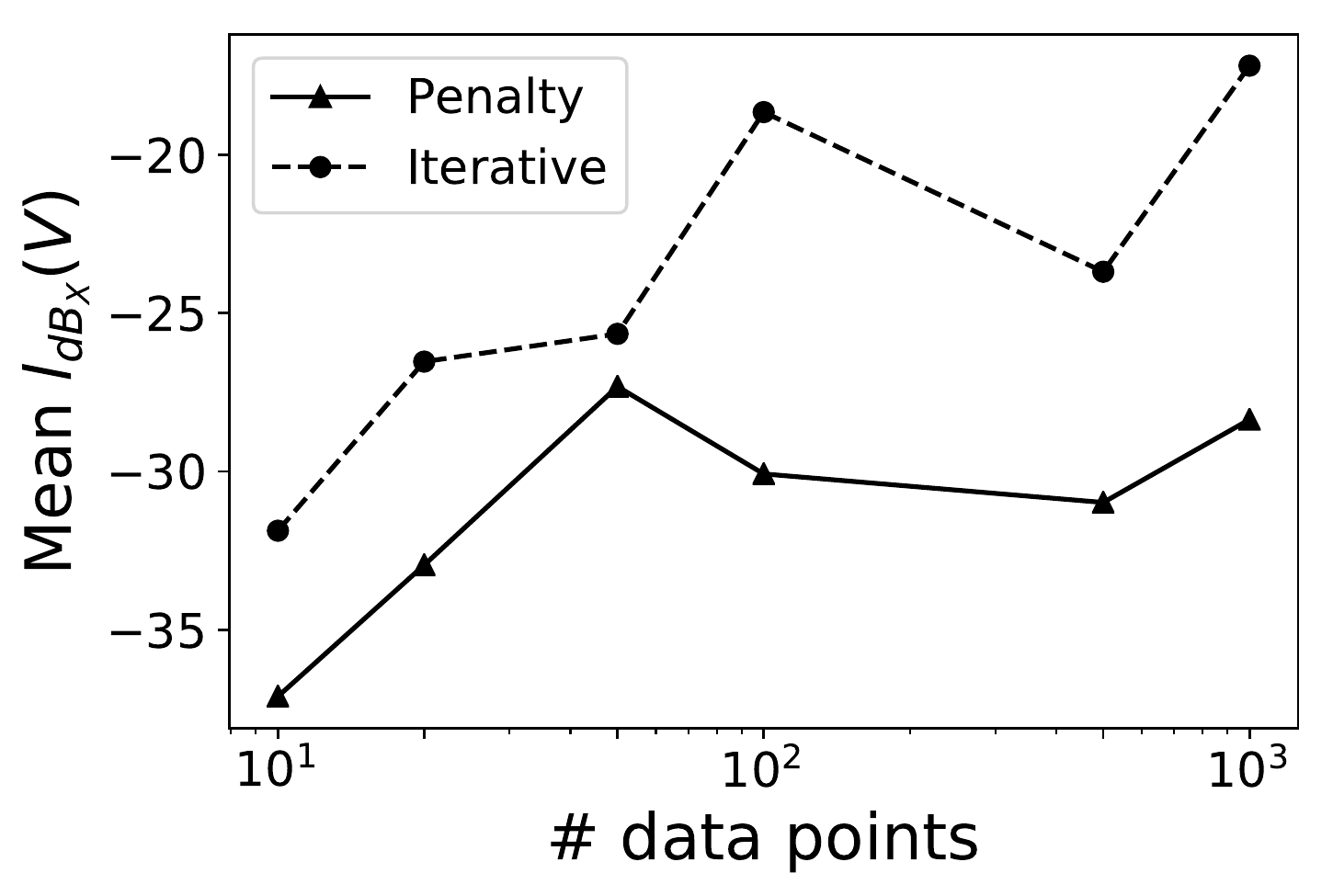}}%
\caption[A set of four subfigures.]{{Effect of the number of data points on the mean values of ASR, SNR and $l_{\text{dB}_{\mathbf x}}(\mathbf v)$ for the test set.
Top row (a, b, c): Targeted attack on the 1D CNN Gamma model \cite{abdoli2019end}. Bottom row (d, e, f): Untargeted attack on the SincNet model~\cite{ravanelli2018speaker}.}
}
\label{fig:datapoints}%
\end{figure*}
\begin{center}
\begin{table*}[ht]
\caption{Results of the untargeted attack in terms of mean values of ASR, SNR and $l_{\text{dB}_{\mathbf x}}(\mathbf v)$ on the test set. The perturbation is generated by a single randomly selected sample of each class. The proposed penalty method is used to generate attacks against the 1D CNN Gamma model \cite{abdoli2019end}.}
\centering
\begin{tabular}{lrrrrrrrrrrr}
\toprule
 & \multicolumn{10}{c}{Available Class} &  \\ \cmidrule(l){2-12}
 & \multicolumn{1}{c}{AI} & \multicolumn{1}{c}{CA} & \multicolumn{1}{c}{CH} & \multicolumn{1}{c}{DO} & \multicolumn{1}{c}{DR} & \multicolumn{1}{c}{EN} & \multicolumn{1}{c}{GU} & \multicolumn{1}{c}{JA} & \multicolumn{1}{c}{SI} & \multicolumn{1}{c}{ST} & \multicolumn{1}{c}{Mean} \\ \midrule
ASR  & 0.641 & 0.722 & 0.730 & 0.732 & 0.720 & 0.561 & 0.688 & 0.796 & 0.677 & 0.713 & 0.698\\
SNR  & 17.718 & 18.317 & 19.597 & 18.813 & 16.699 & 19.039 & 17.908 & 20.046 &  18.014 & 18.743 & 18.489 \\
$l_{\text{dB}_{\mathbf x}}(\mathbf v)$ & -15.130 & -16.574 & -16.322 & -16.052 & -15.048 & -16.963 & -15.672 & -15.510 & -15.088 & -16.086 & -15.844 \\
\bottomrule
\end{tabular}
\label{tab:one_spl}
\end{table*}
\end{center}
We now consider the influence of the number of training data points on the quality of the universal perturbations. Fig.~\ref{fig:datapoints} shows the ASR and mean SNR achieved on the test set with different number of data points, considering two target models. The untargeted attack is evaluated on SincNet and the targeted attack is evaluated on 1D CNN Gamma model. For targeted attacks, the target class is \textit{"Gun shot"}. For both targeted and untargeted scenarios, penalty method produces better ASR when the perturbations are crafted with a lower number of data points. For the untargeted scenario, iterative method produces perturbations with a slightly better mean SNR than those produced by the penalty method. However, this difference is perceptually negligible. When the number of data points are limited, the penalty method also produces better results in terms of mean $l_{\text{dB}_{\mathbf x}}(\mathbf v)$. However, the iterative method produces slightly better results in terms of mean $l_{\text{dB}_{\mathbf x}}(\mathbf v)$ when more data points are available (e.g. more than 50 samples). For the targeted attack scenario, when the number of data points are limited (e.g. lower than 100), the iterative method also produces perturbations with a slightly better mean SNR. However, when the number of data points increases, the penalty method produces better perturbations in terms of SNR than those produced by the iterative method. For such a scenario, the penalty method also produces quieter perturbations in terms of mean $l_{\text{dB}_{\mathbf x}}(\mathbf v)$. The Greedy algorithm used in the iterative method is designed to generate perturbations with the least possible power level. At each iteration, if the perturbation vector misclassifies the example, it will be ignored for the next iterations. Therefore, it is impossible for the attacker to obtain higher ASRs at the expense of having a universal perturbation with slightly higher SPL, especially when the number of samples is limited. However, in the penalty method, the algorithm is able to generate more successful universal perturbations at the expense of having a universal perturbation with a negligible higher power level as long as the gradient-based algorithm can minimize the objective functions defined in Eqs.~\eqref{eq:penalty_targeted} and~\eqref{eq:penalty_untargeted}. Moreover, for all iterations of the penalty method, the algorithm exploits all available data to minimize the objective function for generating the universal perturbation.

Another advantage of the proposed penalty method is that it is also able to generate perturbations when a single audio example is available to the attacker. For such an aim, a single audio sample of each class is randomly selected from the training set, and the objective functions defined in Eqs.~\eqref{eq:penalty_targeted} and \eqref{eq:penalty_untargeted} are minimized iteratively using Algorithm~\ref{Univ_pen} for targeted and untargeted attacks, respectively. For this experiment, we set the penalty coefficient $c$=0.2 and the confidence value $\kappa$=90 for both untargeted and targeted attack scenarios. The algorithm is executed for 19 iterations and the perturbation produced is used to perturb all audio samples of the test set, which are then used to fool the 1D CNN Gamma model.

\begin{figure*}[ht]%
\centering
\subfigure[]{%
\includegraphics[width=.315\textwidth]{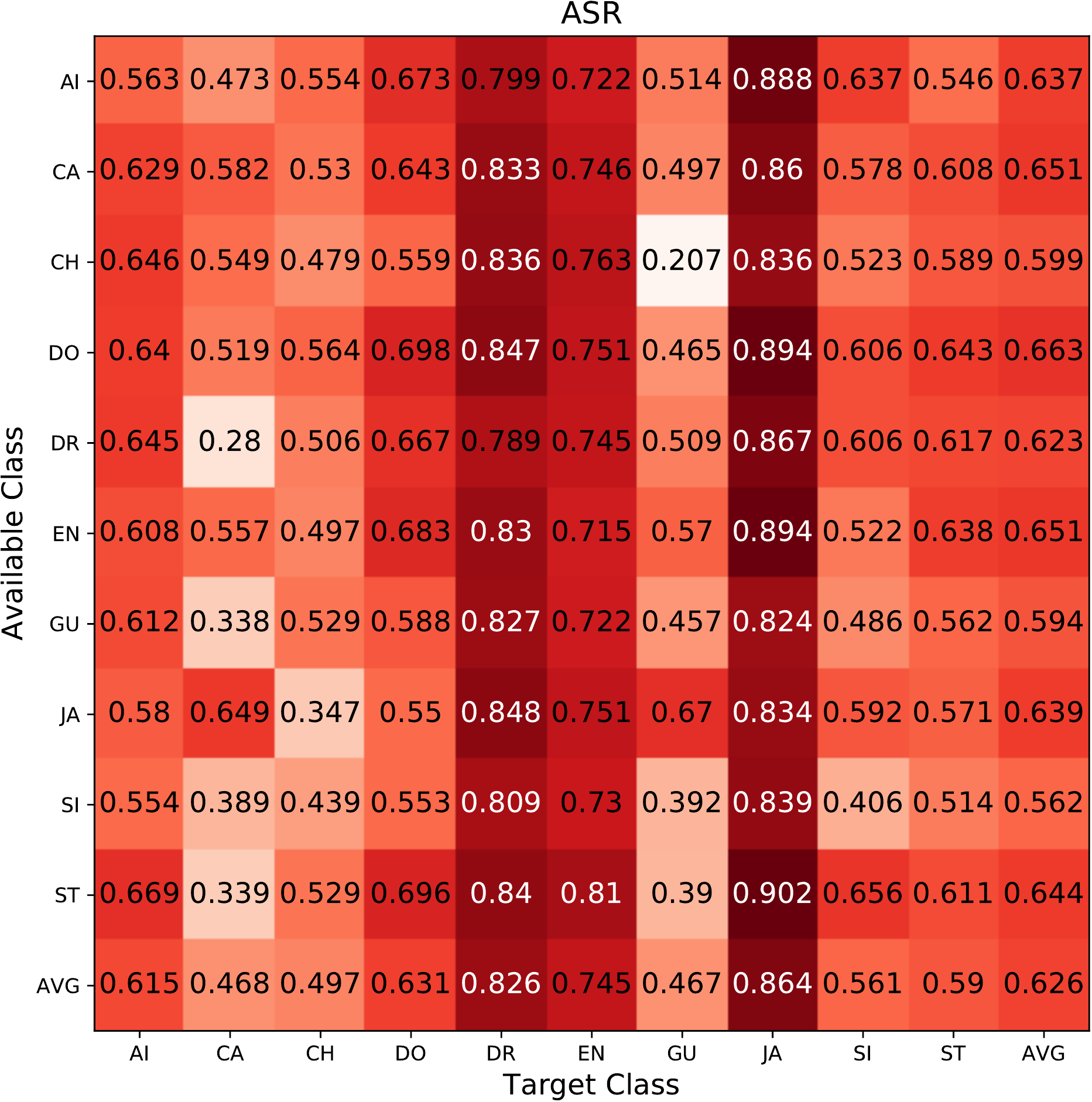}}%
\hspace{1em}
\subfigure[]{%
\includegraphics[width=.315\textwidth]{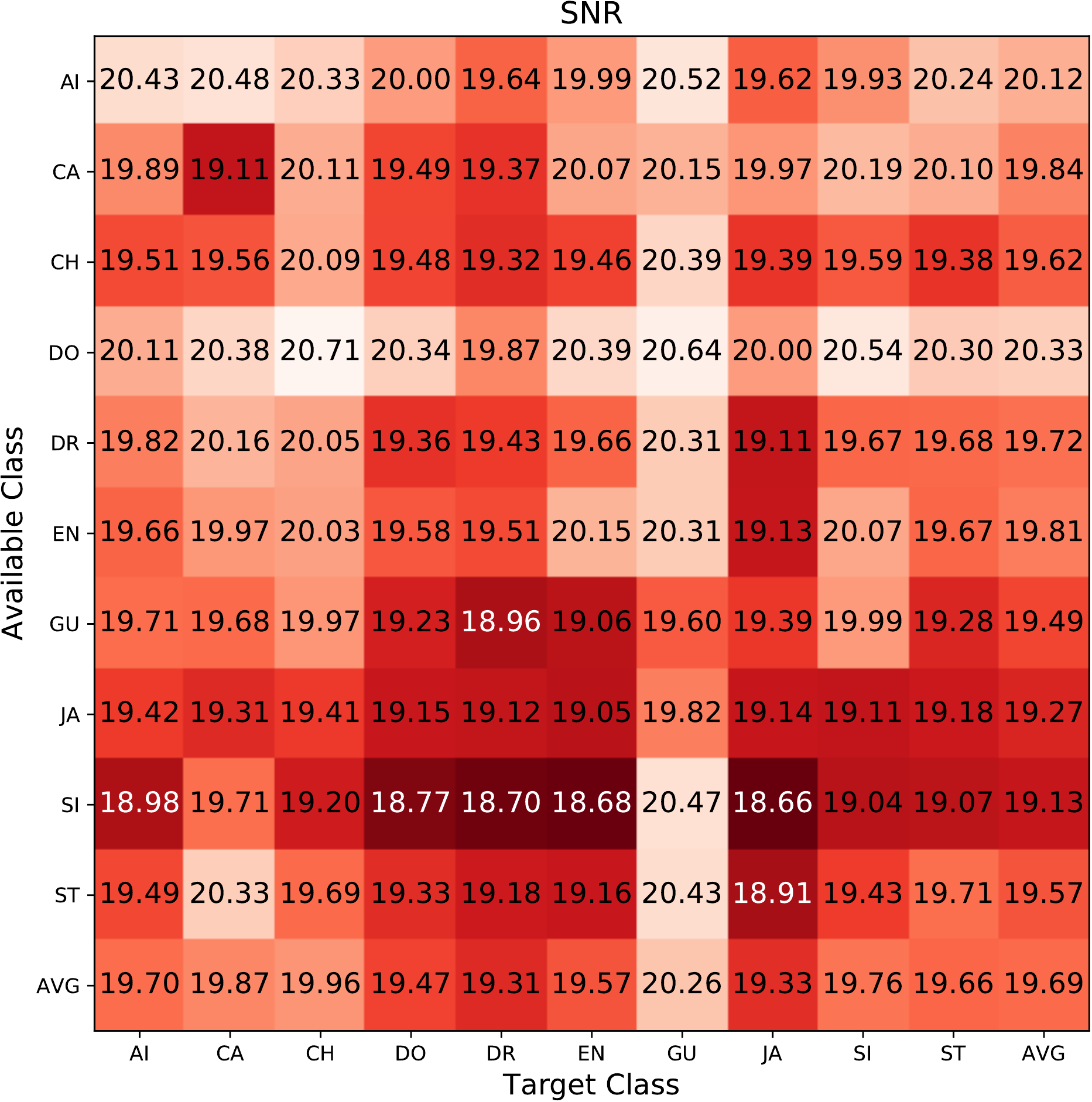}}%
\hspace{1em}
\subfigure[]{%
\includegraphics[width=.315\textwidth]{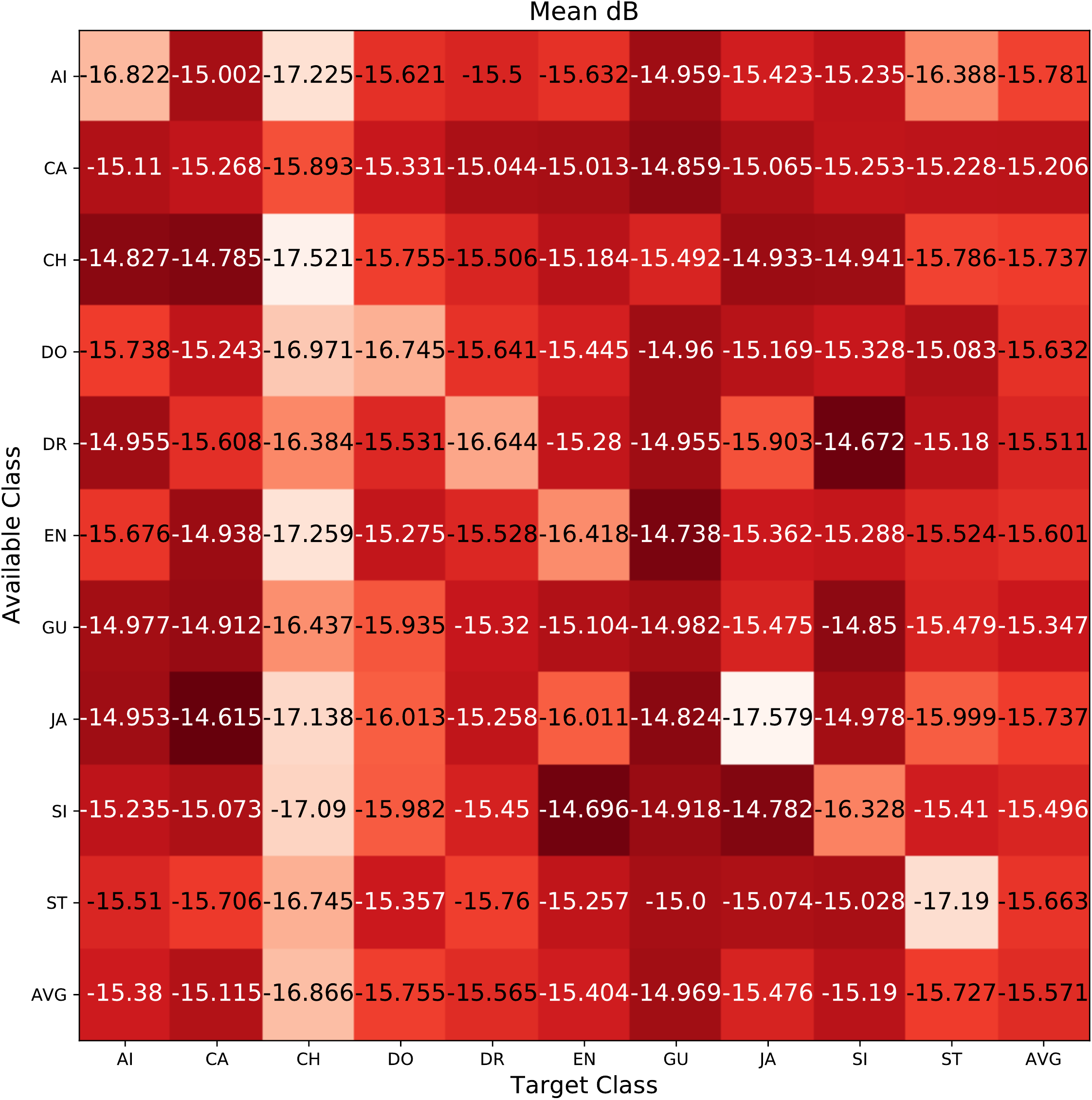}}
\caption{Targeted mean values of ASR (a), SNR (b) and $l_{\text{dB}_{\mathbf x}}(\mathbf v)$ on the test set. Proposed penalty method is used for crafting the perturbation. The perturbation is generated by a single randomly selected sample of each class in order to attack the 1D CNN Gamma model \cite{abdoli2019end} for a specific target class.}
\label{fig:one_spl}
\end{figure*}

Table~\ref{tab:one_spl} shows the results of the untargeted attack in terms of ASR and SNR on the perturbed samples of the test set. Fig.~\ref{fig:one_spl} shows the results of the targeted attack in terms of SNR, ASR and mean $l_{\text{dB}_{\mathbf x}}(\mathbf v)$, considering all classes. In this case, the perturbation is also generated from a single randomly selected audio sample of each class in order to attack the model for a specific target class. Mean ASR of 0.698 and 0.602 are achieved for untargeted and targeted attack scenarios, respectively. Moreover, mean SNR of 18.489 dB and 19.690 dB are also achieved for untargeted and targeted attack scenarios, respectively. Mean $l_{\text{dB}_{\mathbf x}}(\mathbf v)$ of -15.844 dB and 15.571 dB are also achieved for untargeted and targeted attack scenarios, respectively. Similar results were also obtained for all other target models.
\begin{center}
\begin{table*}[htpb!]
\centering
\caption{Transferability of adversarial samples generated by both iterative and penalty methods between pairs of the models for untargeted and targeted attacking scenarios. The cell $(i, j)$ indicates the ASR of the adversarial sample generated for model $i$ (row) evaluated over model $j$ (column)}.

Panel (A): Untargeted attack\\~\\

\begin{tabular}{clccccc}

\toprule

Method                     & Model         & 1D CNN Rand  & 1D CNN Gamma  & ENVnet-V2  & SincNet  & SincNet+VGG19   \\ 
\midrule
\multirow{5}{*}{Iterative} & 1D CNN Rand   & N/A	& 0.371  & 0.175  & 0.300 & 0.200 \\
                           & 1D CNN Gamma  & 0.340  & N/A    & 0.245  & 0.627 & 0.530 \\
                           & ENVnet-V2     & 0.269  & 0.339  & N/A    & 0.304 & 0.310 \\
                           & SincNet       & 0.164  & 0.207  & 0.152  & N/A   & 0.237 \\
                           & SincNet+VGG19 & 0.176  & 0.191  & 0.134  & 0.228 & N/A   \\ \midrule
\multirow{5}{*}{Penalty}   & 1D CNN Rand   & N/A   & 0.558 & 0.342 & 0.675 & 0.388   \\
                           & 1D CNN Gamma  & 0.423 & N/A   & 0.295 & 0.698 & 0.672   \\
                           & ENVnet-V2     & 0.390 & 0.503 & N/A   & 0.490 & 0.479    \\
                           & SincNet       & 0.142 & 0.170 & 0.103 & N/A   & 0.185    \\
                           & SincNet+VGG19 & 0.253 & 0.192 & 0.194 & 0.340 & N/A  \\
\bottomrule\\
\end{tabular}

Panel (B): Targeted attack\\~\\

\begin{tabular}{clccccc}
\toprule
Method                     & Model         & 1D CNN Rand  & 1D CNN Gamma  & ENVnet-V2  & SincNet  & SincNet+VGG19   \\ 
\midrule

\multirow{5}{*}{Iterative} & 1D CNN Rand   & N/A   & 0.352 & 0.186  & 0.534  & 0.232  \\
                           & 1D CNN Gamma  & 0.285 & N/A   & 0.210  & 0.476  & 0.423     \\
                           & ENVnet-V2     & 0.292 & 0.340 & N/A    & 0.327  & 0.356 	   \\
                           & SincNet       & 0.126 & 0.131 & 0.105  & N/A    & 0.215 \\
                           & SincNet+VGG19 & 0.185 & 0.149 & 0.172  & 0.237  & N/A \\ \midrule
\multirow{5}{*}{Penalty}   & 1D CNN Rand   & N/A   & 0.119 & 0.192 & 0.130 & 0.109     \\
                           & 1D CNN Gamma  & 0.130 & N/A   & 0.117 & 0.251 & 0.134	     \\
                           & ENVnet-V2     & 0.127 & 0.153 & N/A   & 0.146 & 0.159     \\
                           & SincNet       & 0.103 & 0.106 & 0.103 & N/A   & 0.129   \\
                           & SincNet+VGG19 & 0.102 & 0.113 & 0.099 & 0.142 & N/A      \\
\bottomrule
\end{tabular}

\label{tab:transferability}
\end{table*}
\end{center}
Finally, Table~\ref{tab:transferability} shows the transferability of perturbed audio samples of the test set of UrbanSound8k dataset generated by one model to other models. Generating transferable adversarial perturbations among the proposed models is a challenging task~\cite{abdullah2020faults,subramanian2019adversarial}. Abdullah \etal\cite{abdullah2020faults} have shown that transferability is challenging for most of the current audio attacks. This problem is still more challenging to gradient-based optimization attacks. Liu \etal\cite{liu2016delving} introduced an ensemble-based approach that seems to be a promising research direction to produce more transferable audio examples.
\section{Conclusion}
\label{sec:conclusion}
In this study, we proposed an iterative method and a penalty method for generating targeted and untargeted universal adversarial audio perturbations. Both methods were used for attacking a diverse family of end-to-end audio classifiers based on deep models and the experimental results have shown that both methods can easily fool all models with a high success rate. It is also proved that the proposed penalty method converges to an optimal solution.  

Although the perturbations produced by both methods can target the models with high success rate, in listening tests humans can detect the additive noise but the adversarial audios can be recognized as their original labels. Developing effective universal adversarial audio examples using the principle of psychoacoustics and auditory masking \cite{qin2019imperceptible, abdullah2020faults} to reduce the level of noise introduced by the adversarial perturbation is our current goal. Moreover, it is also an interesting direction for designing appropriate audio intelligibility metrics to asses the quality of perturbed samples. 

Different from image-based universal adversarial perturbations \cite{Moosavi-Dezfooli_2017_CVPR}, the universal audio perturbations crafted in this study do not perform well in the physical world (played over-the-air). Combining the methods proposed in this paper with recent studies on generating robust adversarial attacks on several transformations on the audio signal \cite{yakura2018robust, qin2019imperceptible} may be a promising way to craft robust physical attacks against audio systems. Moreover, proposing an ensemble-based method for crafting more transferable attacks is also a promising direction for future work \cite{liu2016delving}. 

Finally, proposing a defensive mechanism \cite{esmaeilpour2020class,zeng2019multiversion, esmaeilpour2019robust, yang2018characterizing} against such universal perturbations is also an important issue. Using methods like adversarial training \cite{madry2018towards, sun2018training} against such attacks is also considered as one of our future research directions. Moreover, using the proposed methods for targeting sequence-to-sequence models (\textit{e.g.} speech-to-text) is another interesting research direction.

\appendices
\section{Solving ${\mathbf w}_{i}$ for ${\mathbf v^\prime}$}
\label{app:A_math_step}
\begin{equation*}
{\mathbf w}_{i}=\frac{1}{2}(\tanh ({\mathbf x}_{i}^\prime+{\mathbf v^\prime})+1)\\
\end{equation*}
Substituting ${\mathbf x}_{i}^\prime+{\mathbf v^\prime}$ by ${\mathbf b}_{i}$:
\begin{equation*}
{\mathbf w}_{i}=\frac{1}{2}(\tanh ({\mathbf b}_{i})+1)
\end{equation*}
By using the definition of $\tanh$ function:
\begin{equation*}
\tanh({\mathbf b_{i}})=\frac{-e^{-\mathbf b_{i}}+e^{\mathbf b_{i}}}{e^{-\mathbf b_{i}}+e^{\mathbf b_{i}}},
\end{equation*}
we have:
\begin{equation*}
2{\mathbf w}_{i}-1=\frac{-e^{-\mathbf b_{i}}+e^{\mathbf b_{i}}}{e^{-\mathbf b_{i}}+e^{\mathbf b_{i}}}.
\end{equation*}

Substituting $e^{\mathbf b_{i}}$ by $\mathbf u_{i}$:
\begin{gather*}
2{\mathbf w}_{i}-1=\frac{-\mathbf u_{i}^{-1}+\mathbf u_{i}}{\mathbf u_{i}^{-1}+\mathbf u_{i}}\\[1pt]
\Rightarrow 2{\mathbf w}_{i}-1 = \frac{-1+\mathbf u_{i}^{2}}{1+\mathbf u_{i}^{2}}\\[1pt]
\Rightarrow 2{\mathbf w}_{i}\left(1+\mathbf u_{i}^{2} \right)-\left(1+\mathbf u_{i}^{2} \right)=-1+\mathbf u_{i}^{2}\\[1pt]
\Rightarrow  2{\mathbf w}_{i}+2{\mathbf w}_{i}\mathbf u_{i}^{2}-\mathbf u_{i}^{2}=\mathbf u_{i}^{2}\\[1pt]
\Rightarrow  2{\mathbf w}_{i}=2\mathbf u_{i}^{2}-2{\mathbf w}_{i}\mathbf u_{i}^{2}\\[1pt]
\Rightarrow  {\mathbf w}_{i}=\mathbf u_{i}^{2}\left(1-{\mathbf w}_{i}\right)\\[1pt]
\Rightarrow  \frac{{\mathbf w}_{i}}{1-{\mathbf w}_{i}}=\mathbf u_{i}^{2}
\end{gather*}

$\mathbf u_{i}$ has two solutions:
\begin{align*}
\Rightarrow  \mathbf u_{i}=\pm\sqrt{\frac{{\mathbf w}_{i}}{1-{\mathbf w}_{i}}}
\end{align*}
Considering $e^{\mathbf b_{i}}=\mathbf u_{i}$, for the first solution we have:

\begin{gather*}
e^{\mathbf b_{i}}=\sqrt{\frac{{\mathbf w}_{i}}{1-{\mathbf w}_{i}}}\\[1pt]
\Rightarrow e^{\mathbf b_{i}} = \left (\frac{{\mathbf w}_{i}}{1-{\mathbf w}_{i}}\right )^{\frac{1}{2}}\\[1pt]
\Rightarrow \ln \left (e^{\mathbf b_{i}}\right ) = \ln \left (\frac{{\mathbf w}_{i}}{1-{\mathbf w}_{i}}\right )^{\frac{1}{2}}\\[1pt]
\Rightarrow \mathbf b_{i} =  \frac{1}{2} \ln \left (\frac{{\mathbf w}_{i}}{1-{\mathbf w}_{i}}\right ).
\end{gather*}

By substituting back ${\mathbf b}_{i}$ by ${\mathbf x}_{i}^\prime+{\mathbf v^\prime}$, we have:
\begin{gather*}
\Rightarrow
{\mathbf x}_{i}^\prime+{\mathbf v^\prime} = \frac{1}{2}\ln\left(\frac{{\mathbf w}_{i}}{1-{\mathbf w}_{i}}\right )\\[1pt]
\Rightarrow {\mathbf v^\prime} = \frac{1}{2}\ln\left(\frac{{\mathbf w}_{i}}{1-{\mathbf w}_{i}}\right )-{\mathbf x}_{i}^\prime.
\end{gather*}

For the second solution we have:
\begin{gather*}
\mathbf b_{i} = {\frac{1}{2}} \ln \left (-\frac{{\mathbf w}_{i}}{1-{\mathbf w}_{i}} \right )
\end{gather*}
However, since \(0 \leq {\mathbf w}_{i} \leq 1\) and the natural logarithm of a negative number is undefined,
the second solution for $\mathbf u_{i}$ is invalid.
\section{Solving $\mathbf v^\prime$ for $\mathbf v$}
\label{app:A_math_step_v^pr}
\begin{equation*}
\mathbf v^{\prime}=\arctanh\left ((2\mathbf v-1)*(1-\epsilon)\right).\\
\end{equation*}
Substituting $(2\mathbf v-1)*(1-\epsilon)$ by $\mathbf s$:
\begin{gather*}
\mathbf v^{\prime}=\arctanh\left (\mathbf s\right)\\
\Rightarrow
\tanh(\mathbf v^{\prime}) = \tanh (\arctanh (\mathbf s))\\
\Rightarrow
\tanh(\mathbf v^{\prime}) = \mathbf s.
\end{gather*}
By substituting back $\mathbf s$ by $(2\mathbf v-1)*(1-\epsilon)$ we have:
\begin{gather*}
\tanh(\mathbf v^{\prime}) = (2\mathbf v-1)*(1-\epsilon)\\
\Rightarrow
\tanh(\mathbf v^{\prime}) = 2\mathbf v - 2\mathbf v\epsilon -1 + \epsilon\\
\Rightarrow
\tanh(\mathbf v^{\prime}) = \mathbf v (2-2\epsilon) -1 + \epsilon\\
\frac{\tanh(\mathbf v^{\prime})+1-\epsilon}{2-2\epsilon} = \mathbf v
\end{gather*}


\section{Statistical Test}
\label{app:Statistical_test}
Since the ASRs reported in this paper are the proportion of the successfully attacked samples by the two methods (iterative and penalty), the test of hypothesis concerning two proportions can be applied here \cite{johnson2017miller}. Compare the ASRs obtained by two proposed methods (iterative and penalty) presented in Table \ref{tab:fooling_ratio_all} and Table \ref{tabl:fooling_ratio_all_speech} and let $\widehat{p_{l}}$ and $\widehat{p_{h}}$ be the ASR of the method with lower ASR and the ASR of the method with higher ASR, respectively. The statistical test \cite{johnson2017miller} is given by :
\begin{gather*}
Z= \frac{\widehat{p_{l}} - \widehat{p_{h}}}{\sqrt{2\widehat{p}(1-\hat{p})/m}}, \quad  \widehat{p} = \frac{(\widehat{X_{l}}+\widehat{X_{h}})}{2m}.
\end{gather*}
Where $\widehat{X_{l}}$ and $\widehat{X_{h}}$ are the number of successfully attacked samples by the method with lower ASR and by the method with higher ASR, respectively. Parameter $m$ is also the number of samples in the test set. The intention is to prove the ASR of the two methods \textit{i.e.}, we want to establish that: $\widehat{p_{l}}<\widehat{p_{h}}$ so,
\begin{itemize}
  \item $H_{0}: \widehat{p_{l}}=\widehat{p_{h}}$ (Null hypothesis) 
  \item $H_{a}: \widehat{p_{l}}<\widehat{p_{h}}$ (Alternative hypothesis).
\end{itemize}
The null hypothesis is rejected if $Z<-z_{\alpha}$, where $z_{\alpha}$ is obtained from a standard normal distribution that is related to level of significance, $\alpha$. If the condition is true we can reject $H_{0}$ and accept $H_{a}$. Table~\ref{tab:statistical_test} shows $\widehat{p_{l}}$ and $\widehat{p_{h}}$ for all of the target models for each attacking scenario as well as the corresponding $Z$ values. The ASR of the method which produces higher ASR is considered as $\widehat{X_{h}}$ and vice versa. From the standard normal distribution function \cite{johnson2017miller}, we know that $z_{0.057}=1.58$. Since all of the $Z$ values in Table~\ref{tab:statistical_test} are below -1.58 ($Z<-z_{0.057}$), we can reject the null hypothesis at least with 94.3\% significance level (1-$\alpha$) for all of the target models. In other words, the method corresponding to ASR of $\widehat{p_{h}}$ is more effective with 94.3\% significance level. For most of the models, the attacking method with higher ASR, is more effective with much higher significance level. We can conclude that results are statistically significant.

\begin{center}
\begin{table}[ht]
\centering
\caption{Statistical test of the two attacking methods. $\widehat{p_{l}}$ and $\widehat{p_{h}}$ for all of the target models for each attacking scenario as well as the corresponding $Z$ values. $\widehat{p_{l}}$ and $\widehat{p_{h}}$ are derived from ASRs reported in Table \ref{tab:fooling_ratio_all} and Table \ref{tabl:fooling_ratio_all_speech}}
\begin{tabular}{clccc}
\toprule
Attacking sc.               & Model         & $\widehat{p_{l}}$  & $\widehat{p_{h}}$  & $Z$     \\ 
\midrule
\multirow{5}{*}{Targeted} & 1D CNN Rand    & 0.672	& 0.854 & -8.946  \\
                           & 1D CNN Gamma  & 0.795	& 0.888	& -5.323	  \\
                           & ENVnet-V2     & 0.767	& 0.877	& -6.011	 \\
                           & SincNet       & 0.899	& 0.971	& -6.105	\\
                           & SincNet+VGG19 & 0.872	& 0.898	& -1.703	 \\
                           & SpchCMD       & 0.850	& 0.855	& -3.969	 \\ \midrule
\multirow{5}{*}{Untargeted}& 1D CNN Rand   & 0.412 & 0.876	& -20.257	     \\
                           & 1D CNN Gamma  & 0.737 & 0.858	& -6.294	    \\
                           & ENVnet-V2     & 0.669	& 0.831	& -7.820	    \\
                           & SincNet       & 0.886	& 0.919	& -2.325	    \\
                           & SincNet+VGG19 & 0.838	& 0.865	& -1.587	   \\
                           & SpchCMD       & 0.834	& 0.875	& -32.737	 \\ 
\bottomrule
\end{tabular}
\label{tab:statistical_test}
\end{table}
\end{center}

\section{Detailed targeted attack results}
\label{app:Detailed}
Tables~\ref{tab:detailed_1D CNN Rand} to~\ref{tab:detailed_SpchCMD} show the detailed mean vlaues of ASR, SNR and $l_{\text{dB}_{\mathbf x}}(\mathbf v)$ on the target models in the targeted attack scenario for training and test sets. For Tables~\ref{tab:detailed_1D CNN Rand} to~\ref{tab:detailed_SincNet+VGG19} the measures are reported for each specific target class of UrbanSound8k \cite{Salamon:2014:DTU:2647868.2655045} and for the Table~\ref{tab:detailed_SpchCMD}, the measures are reported for each specific target class of speech commands dataset \cite{warden2018speech}. The target classes of Urbansound8K dataset \cite{Salamon:2014:DTU:2647868.2655045} are: Air conditioner (AI), Car horn (CA), Children playing (CH), Dog bark (DO), Drilling (DR), Engine (EN) idling, Gun shot (GU), Jackhammer (JA), Siren (SI), Street music (ST).

\begin{center}
\begin{table*}[htpb!]
\caption{Mean values of ASR, SNR and $l_{\text{dB}_{\mathbf x}}(\mathbf v)$ for targeting each label of UrbanSound8k \cite{Salamon:2014:DTU:2647868.2655045} dataset. The target model is 1D CNN Rand. Higher ASRs are in boldface.}
\centering
\begin{tabular}{llllllllllll}
\toprule
     &        & \multicolumn{10}{c}{Target Classes}                     \\ \cline{3-12}
Method    &        & AI  & CA  & CH  & DO  & DR  & EN  & GU  & JA  & SI  & ST  \\
\midrule
\multirow{4}{*}{Iterative} & ASR training set & 0.962 &	0.900 & 0.924 & 0.920 & 0.941 & 0.937 & 0.919 & 0.907 & 0.936 & 0.909  \\
 & ASR test set  & 0.636 & 0.741 & 0.602 & 0.666 & 0.716 & 0.656 & 0.808 & 0.662 & 0.640 & 0.593 \\
 & SNR (dB) test set & 25.396 & 24.234 & 25.751 & 25.256 & 26.494 & 25.715 & 24.689 & 25.399 & 25.623 & 24.651 \\
 & $l_{\text{dB}_{\mathbf x}}(\mathbf v)$ & -18.416 & -18.416 & -18.416 &	-18.416 & -18.416 &	-18.416 & -18.416 &	-18.416 & -18.416 & -18.416 \\
\midrule 
\multirow{4}{*}{Penalty} & ASR training set & 0.907 & 0.917 & 0.916 & 0.919 & 0.906 & 0.932 & 0.929 & 0.921 & 0.906 & 0.917  \\
 & ASR test set  & \textbf{0.846} & \textbf{0.872} & \textbf{0.807} & \textbf{0.832} & \textbf{0.860} & \textbf{0.890} & \textbf{0.905} & \textbf{0.871} & \textbf{0.822} & \textbf{0.834}  \\
 & SNR (dB) test set & 24.170 & 22.492 & 23.239 & 22.532 & 24.371 & 23.647 & 24.688 & 23.445 & 23.172 & 22.920  \\
 & $l_{\text{dB}_{\mathbf x}}(\mathbf v)$ & -18.848 & -15.553 & -15.421 &	-14.980 & -19.406 &	-18.046 & -16.627 &	-16.752 & -16.394 &	-15.590\\
\bottomrule
\end{tabular}
\label{tab:detailed_1D CNN Rand}
\end{table*}
\end{center}
\begin{center}
\begin{table*}[ht]
\caption{Mean values of ASR, SNR and $l_{\text{dB}_{\mathbf x}}(\mathbf v)$ for targeting each label of UrbanSound8k \cite{Salamon:2014:DTU:2647868.2655045} dataset. The target model is 1D CNN Gamma. Higher ASRs are in boldface.}
\centering
\begin{tabular}{llllllllllll}
\toprule
 &    & \multicolumn{10}{c}{Target Classes}   \\ \cline{3-12} 
 Method & & AI  & CA  & CH  & DO  & DR  & EN  & GU  & JA  & SI  & ST  \\
\midrule
\multirow{4}{*}{Iterative} & ASR training set & 0.905 &	0.938 &	0.952 &	0.941 &	0.968 &	0.936 &	0.959 &	0.974 &	0.941 &	0.934  \\
 & ASR test set  & 0.745 &	0.874 &	0.744 &	0.810 &	0.815 &	0.746 &	0.887 &	0.795 &	0.772 & 0.761 \\
 & SNR (dB) test set & 21.766 &	23.091 & 23.384 & 24.829 & 25.339 &	22.620 & 24.734 & 24.530 & 22.519 & 23.058 \\
 & $l_{\text{dB}_{\mathbf x}}(\mathbf v)$ & -18.416 & -18.416 & -18.416 &	-18.416 & -18.416 &	-18.416 & -18.416 &	-18.416 & -18.416 & -18.416 \\
\midrule 
\multirow{4}{*}{Penalty} & ASR training set & 0.916 & 0.928 & 0.923 & 0.912 & 0.909 & 0.902 & 0.926 & 0.910 & 0.903 & 0.900  \\
 & ASR test set  & \textbf{0.871} & \textbf{0.920} & \textbf{0.895} & \textbf{0.891} & \textbf{0.879} & \textbf{0.886} & \textbf{0.900} & \textbf{0.899} & \textbf{0.883} & \textbf{0.855}  \\
 & SNR (dB) test set & 21.351 & 21.730 & 22.392 & 22.648 & 24.257 & 22.141 & 23.380 & 24.663 & 21.401 & 22.902 \\
 & $l_{\text{dB}_{\mathbf x}}(\mathbf v)$ & -15.560 &	-15.113 & -16.543 &	-16.950 & -21.505 &	-17.312 & -15.828 &	-21.380 & -16.329 &	-17.229\\
\bottomrule 
\end{tabular}
\label{tab:detailed_1D CNN Gamma}
\end{table*}
\end{center}

\begin{center}
\begin{table*}[ht]
\caption{Mean values of ASR, SNR and $l_{\text{dB}_{\mathbf x}}(\mathbf v)$ for targeting each label of UrbanSound8k \cite{Salamon:2014:DTU:2647868.2655045} dataset. The target model is ENVnet-V2. Higher ASRs are in boldface.}
\centering
\begin{tabular}{llllllllllll}
\toprule
 &    & \multicolumn{10}{c}{Target Classes}   \\ \cline{3-12}
Method   &    & AI  & CA  & CH  & DO  & DR  & EN  & GU  & JA  & SI  & ST  \\
\midrule  
\multirow{4}{*}{Iterative} & ASR training set & 0.939 & 0.928 & 0.921 & 0.925 & 0.882 & 0.923 & 0.902 & 0.929 & 0.901 & 0.906  \\
 & ASR test set  & 0.797 & 0.835 & 0.697 & 0.767 & 0.764 & 0.763 & 0.804 & 0.744 & 0.747 & 0.754  \\
 & SNR (dB) test set & 22.852 & 22.932 & 22.371 & 23.183 & 22.702 & 23.387 & 20.868 & 23.355 & 23.049 & 22.637 \\
 & $l_{\text{dB}_{\mathbf x}}(\mathbf v)$ & -18.416 & -18.416 & -18.416 &	-18.416 & -18.416 &	-18.416 & -18.416 &	-18.416 & -18.416 & -18.416 \\
\midrule 
\multirow{4}{*}{Penalty}   & ASR training set & 0.926 & 0.935 & 0.916 & 0.928 & 0.923 & 0.904 & 0.929 & 0.904 & 0.936 & 0.919  \\
 & ASR test set  & \textbf{0.873} & \textbf{0.902} & \textbf{0.860} & \textbf{0.873} & \textbf{0.867} & \textbf{0.888} & \textbf{0.895} & \textbf{0.879} & \textbf{0.866} & \textbf{0.863} \\
 & SNR (dB) test set & 22.143 & 21.208 & 22.241 & 21.601 & 22.251 & 21.917 & 20.798 & 22.367 & 21.971 & 21.818   \\
 & $l_{\text{dB}_{\mathbf x}}(\mathbf v)$ & -13.571 &	-14.281 & -13.874 &	-12.977 & -15.444 &	-14.541 & -12.176 &	-15.025 & -15.190 & -14.896\\
\bottomrule 
\end{tabular}
\label{tab:detailed_ENVnet-V2}
\end{table*}
\end{center}
\begin{center}
\begin{table*}[ht]
\caption{Mean values of ASR, SNR and $l_{\text{dB}_{\mathbf x}}(\mathbf v)$ for targeting each label of UrbanSound8k \cite{Salamon:2014:DTU:2647868.2655045} dataset. The target model is SincNet. Higher ASRs are in boldface.}
\centering
\begin{tabular}{llllllllllll}
\toprule
 &    & \multicolumn{10}{c}{Target Classes}   \\ \cline{3-12}
Method  &    & AI  & CA  & CH  & DO  & DR  & EN  & GU  & JA  & SI  & ST  \\
\midrule
\multirow{4}{*}{Iterative} & ASR training set & 1.000 & 0.998 & 1.000 & 1.000 & 1.000 & 1.000 & 1.000 & 1.000 & 1.000 & 1.000  \\

 & ASR test set  & 0.754 & 0.955 & 0.931 & 0.854 & 0.916 & 0.919 & \textbf{0.959} & 0.878 & 0.905 & 0.920  \\
 & SNR (dB) test set & 28.852 & 25.245 & 29.940 & 30.968 & 29.004 & 28.910 & 26.639 & 26.966 & 30.356 & 29.800 \\
 & Mean $l_{\text{dB}_{\mathbf x}}(\mathbf v)$ & -20.1006 & -18.416 &	-23.405 &	-24.369	& -19.411	& -20.230 & -18.416	& -18.416	& -24.250 & -23.427 \\
\midrule
\multirow{4}{*}{Penalty}   & ASR training set & 0.935 & 0.974 & 0.964 & 0.948 & 0.985 & 0.994 & 0.924 & 0.970 & 0.966 & 0.961  \\
 & ASR test set  & \textbf{0.941} & \textbf{0.990} & \textbf{0.974} & \textbf{0.957} & \textbf{0.987} & \textbf{0.994} & 0.943 & \textbf{0.975} & \textbf{0.978} & \textbf{0.975} \\
 & SNR (dB) test set & 33.329 & 25.554 & 32.919 & 32.652 & 28.420 & 28.579 & 28.437 & 28.946 & 32.663 & 32.616 \\
 & Mean $l_{\text{dB}_{\mathbf x}}(\mathbf v)$ & -35.161 & -25.740 & -35.174 & -35.117 & -29.397 & -29.583 & -29.214 & -29.408 & -35.149 & -35.219 \\
\bottomrule 
\end{tabular}
\label{tab:detailed_SincNet}
\end{table*}
\end{center}
\begin{center}
\begin{table*}[ht]
\caption{Mean values of ASR, SNR and $l_{\text{dB}_{\mathbf x}}(\mathbf v)$ for targeting each label of UrbanSound8k \cite{Salamon:2014:DTU:2647868.2655045} dataset. The target model is SincNet+VGG. Higher ASRs are in boldface.}
\centering
\begin{tabular}{llllllllllll}
\bottomrule
 &    & \multicolumn{10}{c}{Target Classes}   \\ \cline{3-12}
Method  &    & AI  & CA  & CH  & DO  & DR  & EN  & GU  & JA  & SI  & ST  \\ \midrule
\multirow{3}{*}{Iterative} & ASR training set & 0.990 & 0.992 & 0.993 & 0.994 & 0.995 & 0.972 & 0.925 & 0.996 & 0.999 & 0.994  \\
 & ASR test set  & 0.876 & \textbf{0.918} & 0.855 & 0.887 & 0.863 & 0.831 & \textbf{0.899} & 0.852 & 0.864 & 0.879 \\
 & SNR (dB) test set & 25.571 & 26.434 & 27.734 & 25.674 & 28.890 & 24.276 & 22.930 & 26.621 & 27.292 & 26.607 \\
 & $l_{\text{dB}_{\mathbf x}}(\mathbf v)$ & -18.416 & -18.416 & -18.416 &	-18.416 & -18.416 &	-18.416 & -18.416 &	-18.416 & -18.416 & -18.416 \\
\midrule
\multirow{3}{*}{Penalty}   & ASR training set & 0.902 & 0.908 & 0.903 & 0.922 & 0.921 & 0.903 & 0.920 & 0.922 & 0.924 & 0.935  \\
 & ASR test set  & \textbf{0.899} & 0.889 & \textbf{0.898} & \textbf{0.897} & \textbf{0.897} & \textbf{0.881} & 0.882 & \textbf{0.907} & \textbf{0.919} & \textbf{0.914} \\
 & SNR (dB) test set & 26.417 & 27.191 & 28.784 & 27.085 & 28.448 & 24.626 & 22.411 & 27.330 & 27.276 & 27.787 \\
 & $l_{\text{dB}_{\mathbf x}}(\mathbf v)$ & -20.024 &	-21.601 & -23.594 &	-21.293 & -23.944 &	-18.213 & -14.776 &	-22.494	& -22.045 &	-22.607\\
\bottomrule
\end{tabular}
\label{tab:detailed_SincNet+VGG19}
\end{table*}
\end{center}
\begin{sidewaystable*}[htpb!]
\centering
\caption{Mean values of ASR, SNR and $l_{\text{dB}_{\mathbf x}}(\mathbf v)$ for targeting each label of speech commands dataset \cite{warden2018speech} dataset. The target model is SpchCMD. Higher ASRs are in boldface.}
\centering
\begin{adjustbox}{width=\textwidth}
\begin{tabular}{llllllllllllllllllllllllllllllllll}
\bottomrule
 &    & \multicolumn{32}{c}{Target Classes}   \\ \cline{3-34}
Method  &   & silence & unknown & sheila & nine & stop & bed & four & six & down & bird & marvin & cat & off & right & seven & eight & up & three & happy & go & zero & on & wow & dog & yes & five & one & tree & house & two & left & no \\ \midrule
\multirow{4}{*}{Iterative} & ASR training set & 0.903&	0.902&	0.903&	0.901&	0.908&	0.902&	0.900&	0.902&	0.904&	0.906&	0.901&	0.902&	0.901&	0.902&	0.903&	0.902&	0.902&	0.900&	0.900&	0.901&	0.900&	0.903&	0.900&	0.903&	0.905&	0.904&	0.904&	0.901&	0.905&	0.904&	0.900&	0.900  \\
 & ASR test set & \textbf{0.875}       & 0.893                & \textbf{0.850}       & \textbf{0.868}       & \textbf{0.859}       & \textbf{0.857}       & \textbf{0.871}       & 0.829                & 0.845                & \textbf{0.861}       & 0.841                & \textbf{0.840}       & \textbf{0.843}       & \textbf{0.863}       & 0.838                & 0.872                & \textbf{0.849}       & \textbf{0.871}       & 0.835                & \textbf{0.857}       & \textbf{0.864}       & \textbf{0.863}       & \textbf{0.855}       & \textbf{0.857}       & \textbf{0.854}       & 0.840                & \textbf{0.870}       & 0.838                & \textbf{0.860}       & 0.840                & \textbf{0.856}       & 0.835                \\
 & SNR (dB) test set & 21.559&	27.747&	31.240&	26.656&	32.116&	25.345&	26.726&	29.381&	30.795&	25.127&	26.984&	28.068&	25.640&	27.637&	29.073&	25.217&	25.775&	27.901&	27.404&	28.173&	29.630&	27.074&	26.043&	26.052&	28.267&	26.428&	27.789&	28.529&	26.800&	28.596&	26.782&	26.203 \\
 & $l_{\text{dB}_{\mathbf x}}(\mathbf v)$ & -18.416& 	-18.416&	-20.546&	-18.416&	-22.206&	-18.416&	-18.416&	-19.350&	-22.887&	-18.416&	-18.416&	-18.529&	-18.416&	-18.418&	-20.292&	-18.416&	-18.416&	-19.338&	-18.416&	-18.416&	-19.459&	-18.416&	-18.416&	-18.416&	-18.421&	-18.416&	-18.417&	-18.450&	-18.416&	-18.840&	-18.416&	-18.416 \\
\midrule
\multirow{4}{*}{Penalty}   & ASR training set & 0.900&	0.901&	0.902&	0.905&	0.901&	0.906&	0.903&	0.901&	0.904&	0.900&	0.901&	0.901&	0.901&	0.909&	0.901&	0.900&	0.901&	0.913&	0.900&	0.907&	0.914&	0.901&	0.903&	0.902&	0.907&	0.901&	0.903&	0.903&	0.902&	0.901&	0.903&	0.902  \\
 & ASR test set  & 0.859                & \textbf{0.910}       & 0.849                & 0.867                & 0.858                & 0.835                & 0.838                & \textbf{0.856}       & \textbf{0.864}       & 0.821                & \textbf{0.861}       & 0.830                & 0.832                & 0.837                & \textbf{0.854}       & \textbf{0.883}       & 0.829                & 0.867                & \textbf{0.855}       & 0.848                & 0.837                & 0.857                & 0.823                & 0.825                & 0.843                & \textbf{0.850}       & 0.857                & \textbf{0.849}       & 0.860                & \textbf{0.849}       & 0.851                & \textbf{0.842}       \\
 & SNR (dB) test set & 26.707&	26.693&	26.716&	26.763&	26.714&	26.706&	26.722&	26.728&	26.741&	26.744&	26.729&	26.716&	26.765&	26.720&	26.713&	26.758&	26.729&	26.750&	26.726&	26.738&	26.721&	26.761&	26.712&	26.727&	26.727&	26.735&	26.716&	26.722&	26.728&	26.722&	26.730&	26.769 \\ 
 & $l_{\text{dB}_{\mathbf x}}(\mathbf v)$&  -19.767&	-19.913&	-24.018&	-20.482&	-22.100&	-20.514&	-22.881&	-21.749&	-21.484&	-20.034&	-20.996&	-23.751&	-20.333&	-23.642&	-20.663&	-20.971&	-17.679&	-22.893&	-19.322&	-22.805&	-23.252&	-19.994&	-22.219&	-21.555&	-22.725&	-20.931&	-23.301&	-23.257&	-20.861&	-20.510&	-20.723&	-22.056 \\
\bottomrule
\end{tabular}
\end{adjustbox}
\label{tab:detailed_SpchCMD}
\end{sidewaystable*}

\section{Target models}
\label{app:target_models}
In this study six types of models are targeted. For training all models, categorical crossentropy is used as loss function. Adadelta \cite{zeiler2012adadelta} is used for optimizing the parameters of the models proposed for environmental sound classification. For training the speech command classification model RMSprop is also used as the optimization method. In this section, we present the complete description of the models.

\subsection{1D CNN Rand}
Table \ref{tab:1D-CNN-Rand} shows the configuration of 1D CNN Rand \cite{abdoli2019end}. This model consists of five one-dimensional CLs. The number of kernels of each CL is 16, 32, 64, 128 and 256. The size of the feature maps of each CL is 64, 32, 16, 8 and 4. The first, second and fifth CLs are followed by a one-dimensional max-pooling layer of size of eight, eight and four, respectively. The output of the second pooling layer is used as input to two FC layers on which a drop-out with probability of 0.5 is applied for both layers \cite{srivastava2014dropout}. Relu is used as the activation function for all of the layers. The number of neurons of the FC layers are 128 and 64. In order to reduce the over-fitting, batch normalization is applied after the activation function of each convolution layer \cite{ioffe2015batch}. The output of the last FC layer is used as the input to a softmax layer with ten neurons for classification.

\subsection{1D CNN Gamma}
This model is similar to 1D CNN Rand except that a gammatone filter-bank is used for initialization of the filters of the first layer of this model \cite{abdoli2019end}. Table \ref{tab:1D-CNN-gamma} shows the configuration of this model. The gammatone filters are kept frozen and they are not trained during the backpropagation process. Sixty-four filters are used to decompose the input signal into appropriate frequency bands. This filter-bank covers the frequency range between 100 Hz to 8 kHz. After this layer, batch normalization is also applied \cite{ioffe2015batch}. 

\subsection{ENVnet-V2}
Table \ref{tab:ENVnet-V2} shows the architecture of ENVnet-V2 \cite{tokozume2017learning}. This model extracts short-time frequency features from audio waveforms by using two one-dimensional CLs with 32 and 64 filters, respectively followed by a one-dimensional max-pooling layer. The model then swaps axes and convolves in time and frequency domain features using two two-dimensional CLs each with 32 filters. After the CLs, a two-dimensional max-pooling layer is used. After that, two other two-dimensional CLs followed by a max-pooling layer are used and finally another two-dimensional CL with 128 filters is used. After using two FC layers with 4,096 neurons, a softmax layer is applied for classification. Drop-out with probability of 0.5 is applied to FC layers \cite{srivastava2014dropout}. Relu is used as the activation function for all of the layers. 

\subsection{SincNet}
Table \ref{tab:SincNet} shows the architecture of SincNet \cite{ravanelli2018speaker}. In this model, 80 sinc functions are used as band-pass filters for decomposing the audio signal into appropriate frequency bands. After that, two one-dimensional CLs with 80 and 60 filters are applied. Layer normalization \cite{lei2016layer} is used after each CL. After each CL, max-pooling is used. Two FC layers followed by a softmax layer is used for classification. Drop-out with probability of 0.5 is applied to FC layers \cite{srivastava2014dropout}. Batch normalization \cite{ioffe2015batch} is used after FC layers. In this model, all hidden layers use leaky-ReLU \cite{maas2013rectifier} non-linearity.

\subsection{SincNet+VGG19}
Table \ref{tab:SincNet+VGG19} shows the specification of this architecture. This model uses 227 Sinc filters to extract features from the raw audio signal as it is introduced in SincNet \cite{ravanelli2018speaker}. After applying one-dimensional max-pooling layer of size 218 with stride of one, and layer normalization \cite{lei2016layer}, the output is stacked along time axis to form a 2D representation. This time-frequency representation is used as input to a VGG19 \cite{VGG19} network followed by a FC layer and softmax layer for classification. The parameters of the VGG19 are the same as described in \cite{VGG19} and they are not changed in this study. The output of the VGG19 is used as input of a softmax layer with ten neurons for classification.

\subsection{SpchCMD}
Table~\ref{tab:SpchCMD} shows the architecture of SpchCMD model. The model receives the raw audio input of size of 16,000. The model generates 40 patches of overlapped audio chunks of size of 800. After that, a one-dimensional convolution layer with 64 filters is used. Eleven depth-wise one-dimensional convolution layers with appropriate number of feature maps are then used to extract suitable information from the representations from the last layer. Relu is used as the activation function for all convolution layers. Bach normalization is also used after each convolution layers. After an attention layer and also a global average pooling layer a FC and softmax layer is used to classify the input samples into appropriate classes. This model is trained based on the available recipe from the Github page of the winner of the challenge \footnote{\url{https://github.com/see--/speech\_recognition}}. A batch consists of 384 sample from the dataset is used where up to 40\% of the batch are samples from the test set which are annotated based on pseudo labeling. An ensemble of the three best performed models during the initial experiments on the test is used for the labeling where, the three models predict the same label for the sample. Up to 50\% percent of the samples are also augmented by the use of time shifting with the minimum and maximum range of (-2000, 0) moreover, up to 15\% of the samples are also silent signals.

\begin{center}
\begin{table}[]
\caption{1D CNN Rand architecture.}
\centering
\begin{tabular}{lllll}
\toprule
Layer        & Ksize & Stride & \# of filters & Data shape   \\ \midrule
InputLayer   & -     & -      & -             & (50,999, 1)  \\
Conv1D       & 64    & 2      & 16            & (25,468, 16) \\
MaxPooling1D & 8     & 8      & 16            & (3,183, 16)  \\
Conv1D       & 32    & 2      & 32            & (1,576, 32)  \\
MaxPooling1D & 8     & 8      & 32            & (197, 32)    \\
Conv1D       & 16    & 2      & 64            & (91, 64)     \\
Conv1D       & 8     & 2      & 128           & (42, 128)    \\
Conv1D       & 4     & 2      & 256           & (20, 256)    \\
MaxPooling1D & 4     & 4      & 128           & (5, 256)     \\
FC           & -     & -      & 128           & (128)        \\
FC           & -     & -      & 64            & (64)         \\
FC           & -     & -      & 10            & (10)         \\ \bottomrule
\end{tabular}
\label{tab:1D-CNN-Rand}
\end{table}
\end{center}
\begin{table}[]
\caption{1D CNN Gamma architecture}
\centering
\begin{tabular}{lllll}
\toprule
Layer        & Ksize & Stride & \# of filters & Data shape   \\ \midrule
InputLayer   & -     & -      & -             & (50,999, 1)  \\
Conv1D       & 512   & 1      & 64            & (50,488, 64) \\
MaxPooling1D & 8     & 8      & 64            & (6,311, 64)  \\
Conv1D       & 32    & 2      & 32            & (3,140, 32)  \\
MaxPooling1D & 8     & 8      & 32            & (392, 32)    \\
Conv1D       & 16    & 2      & 64            & (189, 64)    \\
Conv1D       & 8     & 2      & 128           & (91, 128)    \\
Conv1D       & 4     & 2      & 256           & (44, 256)    \\
MaxPooling1D & 4     & 4      & 128           & (11, 256)    \\
FC           & -     & -      & 128           & (128)        \\
FC           & -     & -      & 64            & (64)         \\
FC           & -     & -      & 10            & (10)         \\ \bottomrule
\end{tabular}
\label{tab:1D-CNN-gamma}
\end{table}
\begin{table}[]
\caption{ENVnet-V2 architecture}
\centering
\begin{tabular}{lllll}
\toprule
Layer        & Ksize & Stride & \# of filters & Data shape    \\ \midrule
InputLayer   & -     & -      & -             & (50,999, 1)   \\
Conv1D       & 64    & 2      & 32            & (25,468, 32)  \\
Conv1D       & 16    & 2      & 64            & (12,727, 64)  \\
MaxPooling1D & 64    & 64     & 64            & (198, 64)     \\
swapaxes     & -     & -      & -             & (64, 198, 1)  \\
Conv2D       & (8,8) & (1,1)  & 32            & (57, 191, 32) \\
Conv2D       & (8,8) & (1,1)  & 32            & (50, 184, 32) \\
MaxPooling2D & (5,3) & (5,3)  & 32            & (10, 61, 32)  \\
Conv2D       & (1,4) & (1,1)  & 64            & (10, 58, 64)  \\
Conv2D       & (1,4) & (1,1)  & 64            & (10, 55, 64)  \\
MaxPooling2D & (1,2) & (1,2)  & 64            & (10, 27, 64)   \\
Conv2D       & (1,2) & (1,1)  & 128           & (10, 26, 128)  \\
FC           & -     & -      & 4,096         & (4,096)       \\
FC           & -     & -      & 4,096         & (4,096)        \\
FC           & -     & -      & 10            & (10)          \\ \bottomrule
\end{tabular}
\label{tab:ENVnet-V2}
\end{table}
\begin{table}[]
\caption{SincNet architecture}
\centering
\begin{tabular}{lllll}
\toprule
Layer        & Ksize & Stride & \# of filters & Data shape   \\ \midrule
InputLayer   & -     & -      & -             & (50,999, 1)  \\
SincConv1D   & 251   & 1      & 80            & (50,749, 80) \\
MaxPooling1D & 3     & 1      & 80            & (16,916, 80) \\
Conv1D       & 5     & 1      & 60            & (16,912, 60) \\
MaxPooling1D & 3     & 1      & 60            & (5,637, 60)  \\
Conv1D       & 5     & 1      & 60            & (5,633, 60)  \\
FC           & -     & -      & 128           & (128)        \\
FC           & -     & -      & 64            & (64)         \\
FC           & -     & -      & 10            & (10)         \\ \bottomrule
\end{tabular}
\label{tab:SincNet}
\end{table}
\begin{table}[]
\caption{SincNet+VGG19 architecture}
\centering
\begin{tabular}{lllll}
\toprule
Layer        & Ksize & Stride & \# of filters & Data shape    \\ \midrule
InputLayer   & -     & -      & -             & (50,999, 1)   \\
SincConv1D   & 251   & 1      & 227           & (50,749, 1)   \\
MaxPooling1D & 218   & 1      & 227           & (232, 1)      \\
Reshape      & -     & -      & -             & (232, 227, 1) \\
VGG19 \cite{VGG19}        & -     & -      & -             & (4096)   \\
FC           & -     & -      & 10            & (10)          \\ \bottomrule
\hline
\end{tabular}
\label{tab:SincNet+VGG19}
\end{table}


\begin{table}[]
\caption{SpchCMD architecture}
\centering
\begin{tabular}{lllll}
\toprule
Layer        & Ksize & Stride & \# of filters & Data shape   \\ \midrule
InputLayer          & -     & -      & -             & (16,000)  \\
Time\_slice\_stack  &  40 &    20 &  -         & (800, 40) \\
Conv1D              & 3  &   2   &  64        & (399, 64) \\
DeptwiseConv1D      & 3  &   1   &  128        & (397, 128) \\

DeptwiseConv1D      & 3  &   1   &  192        & (199, 192) \\
DeptwiseConv1D      & 3  &   1   &  192        & (197, 192) \\

DeptwiseConv1D      & 3  &   1   &  256        & (99, 256) \\
DeptwiseConv1D      & 3  &   1   &  256        & (97, 256) \\

DeptwiseConv1D      & 3  &   1   &  320       & (49, 320) \\
DeptwiseConv1D      & 3  &   1   &  320       & (47, 320) \\

DeptwiseConv1D      & 3  &   1   &  384       & (24, 384) \\
DeptwiseConv1D      & 3  &   1   &  384       & (22, 384) \\

DeptwiseConv1D      & 3  &   1   &  448       & (11, 448) \\
DeptwiseConv1D      & 3  &   1   &  448       & (9, 448) \\

AttentionLayer      & -  &   -   &  448       & (9, 448) \\

GolobalAvgPooling   & -  &   -   &  448       & (448) \\
FC                  & -  &   -   &  32        & (32) \\

\bottomrule
\end{tabular}
\label{tab:SpchCMD}
\end{table}

\section{Audio Examples}
\label{app:audio_exampl}
Several randomly chosen examples of perturbed audio samples of Urbansound8k dataset \cite{Salamon:2014:DTU:2647868.2655045} and Speech commands \cite{warden2018speech} datasets are also presented. The audio samples are perturbed based on two presented methods in this study. Targeted and untargeted perturbations are considered. Table \ref{tab:examples_env} shows a list of the samples of audio samples of Urbansound8k dataset \cite{Salamon:2014:DTU:2647868.2655045} and Table \ref{tab:examples_spch} also shows the list of audio samples of Speech Commands \cite{warden2018speech} dataset . Methodology of crafting the samples, target models, SNR of the perturbed samples, $l_{\text{dB}_{\mathbf x}}(\mathbf v)$ of the perturbed samples, detected class of the sample by each model as well as the true class of the samples are also presented. 
\begin{center}
\begin{table*}[htpb!]
\centering
\begin{tabular}{cccccccc}
\toprule
Sample               & Detected Class    & True Class       & Target Model  & Method    & Targeted/Untargeted & SNR & $l_{\text{dB}_{\mathbf x}}(\mathbf v)$ \\ \midrule
DR\_0\_org.wav       & Drilling          & Drilling         & SINCNet       & N/A       & N/A                 & N/A & N/A       \\
DR\_0\_pert\_itr.wav & Gun shot          & Drilling         & SINCNet       & Iterative & Targeted            & 27.025 & -18.416   \\
DR\_0\_pert\_pen.wav & Gun shot          & Drilling         & SINCNet       & penalty   & Targeted            & 28.040 & -29.264   \\
SI\_0\_org.wav       & Siren             & Siren            & SINCNet       & N/A       & N/A                 & N/A    & N/A        \\
SI\_0\_pert\_itr.wav & Airconditioner    & Siren            & SINCNet       & Iterative & Targeted            & 29.244 & -20.101  \\
SI\_0\_pert\_pen.wav & Airconditioner    & Siren            & SINCNet       & penalty   & Targeted            & 33.766 & -35.153  \\
CH\_0\_org.wav       & Children playing  & Children playing & SINCNet       & N/A       & N/A                 & N/A    & N/A    \\
CH\_0\_pert\_itr.wav & Dog bark          & Children playing & SINCNet       & Iterative & Targeted            & 31.784 & -24.369  \\
CH\_0\_pert\_pen.wav & Dog bark          & Children playing & SINCNet       & penalty   & Targeted            & 33.207 & -35.154  \\
ST\_0\_org.wav       & Street music      & Street music     & SINCNet       & N/A       & N/A                 & N/A    & N/A    \\
ST\_0\_pert\_itr.wav & Airconditioner    & Street music     & SINCNet       & Iterative & Targeted            & 28.994 & -20.101  \\
ST\_0\_pert\_pen.wav & Airconditioner    & Street music     & SINCNet       & penalty   & Targeted            & 33.539 & -35.126  \\
DO\_0\_org.wav       & Dog bark          & Dog bark         & SINCNet       & N/A       & N/A                 & N/A    & N/A    \\
DO\_0\_pert\_itr.wav & Drilling          & Dog bark         & SINCNet       & Iterative & Targeted            & 28.861 & -19.410 \\
DO\_0\_pert\_pen.wav & Drilling          & Dog bark         & SINCNet       & penalty   & Targeted            & 28.040 & -29.306   \\
EN\_0\_org.wav       & Engine idling     & Engine idling    & SINCNet+VGG19 & N/A       & N/A                 & N/A    & N/A    \\
EN\_0\_pert\_itr.wav & Drilling          & Engine idling    & SINCNet+VGG19 & Iterative & untargeted          & 26.378 & -14.005  \\
EN\_0\_pert\_pen.wav & Children playing  & Engine idling    & SINCNet+VGG19 & penalty   & untargeted          & 23.444 & -17.651  \\
CA\_0\_org.wav       & Car horn          & Car horn         & SINCNet+VGG19 & N/A       & N/A                 & N/A    & N/A    \\
CA\_0\_pert\_itr.wav & Drilling          & Car horn         & SINCNet+VGG19 & Iterative & untargeted          & 26.175 & -14.005   \\
CA\_0\_pert\_pen     & Street music      & Car horn         & SINCNet+VGG19 & penalty   & untargeted          & 23.341 & -17.588  \\
AI\_0\_org.wav       & Airconditioner    & Airconditioner   & SINCNet+VGG19 & N/A       & N/A                 & N/A    & N/A    \\
AI\_0\_pert\_itr.wav & Jackhammer        & Airconditioner   & SINCNet+VGG19 & Iterative & untargeted          & 25.759 & -14.005  \\
AI\_0\_pert\_pen.wav & Children  playing & Airconditioner   & SINCNet+VGG19 & penalty   & untargeted          & 23.073 & -17.535  \\
SI\_1\_org.wav       & Siren             & Siren            & SINCNet+VGG19 & N/A       & N/A                 & N/A    & N/A    \\
SI\_1\_pert\_itr.wav & Jackhammer        & Siren            & SINCNet+VGG19 & Iterative & untargeted          & 26.334 & -14.005  \\
SI\_1\_pert\_pen.wav & Children  playing & Siren            & SINCNet+VGG19 & penalty   & untargeted          & 23.423 & -17.907 \\
DR\_1\_org.wav       & Drilling          & Drilling         & SINCNet+VGG19 & N/A       & N/A                 & N/A    & N/A     \\
DR\_1\_pert\_itr.wav & Jackhammer        & Drilling         & SINCNet+VGG19 & Iterative & untargeted          & 27.040 & -14.005 \\
DR\_1\_pert\_pen.wav & Children playing  & Drilling         & SINCNet+VGG19 & penalty   & untargeted          & 24.555 & -17.688\\
\bottomrule
\end{tabular}
\caption{List of examples of perturbed audio samples, Methodology of crafting the samples, SNR of the perturbed samples, $dB_{\mathbf x}(\mathbf v)$ of the perturbed samples, target models, and also detected class of the sample by each model and the true class of the samples. The audio files belong to UrbanSound8k dateset \cite{Salamon:2014:DTU:2647868.2655045}. N/A: Not Applicable}
\label{tab:examples_env}

\end{table*}
\end{center}
\begin{center}
\begin{table*}[htpb!]
\centering
\begin{tabular}{ccccccc}
\toprule
Sample                  & Detected Class & True Class        & Method       & Targeted/Untargeted & SNR & $l_{\text{dB}_{\mathbf x}}(\mathbf v)$ \\ \midrule
No\_0\_org.wav          & no             & no         & N/A       & N/A        & N/A    & N/A       \\
No\_0\_pert\_pen.wav    & bed            & no         & penalty   & Targeted   & 29.190 & -20.958   \\
No\_0\_pert\_itr.wav    & bed            & no         & iterative & Targeted   & 27.323 & -18.416   \\
Up\_0\_org.wav          & up             & up         & N/A       & N/A        & N/A    & N/A       \\
Up\_0\_pert\_pen.wav    & unknown        & up         & Penalty   & Targeted   & 24.773 & -20.319   \\
Up\_0\_pert\_itr.wav    & unknown        & up         & iterative & Targeted   & 25.385 & -18.416   \\
Five\_0\_org.wav        & five           & five       & N/A       & N/A        & N/A    & N/A       \\
Five\_0\_pert\_pen.wav  & silence        & five       & penalty   & Targeted   & 25.475 & -20.218   \\
Five\_0\_pert\_itr.wav  & silence        & five       & iterative & Targeted   & 20.293 & -18.416   \\
Down\_0\_org.wav        & down           & down       & N/A       & N/A        & N/A    & N/A       \\
Down\_0\_pert\_pen.wav  & dog            & down       & penalty   & targeted   & 31.342 & -23.030   \\
Down\_0\_pert\_itr.wav  & dog            & down       & iterative & targeted   & 29.130 & -18.416   \\
One\_0\_org.wav         & one            & one        & N/A       & N/A        & N/A    & N/A       \\
One\_0\_pert\_pen.wav   & seven          & one        & penalty   & targeted   & 27.276 & -21.299   \\
One\_0\_pert\_itr.wav   & seven          & one        & iterative & targeted   & 29.351 & -20.292   \\
Right\_0\_org.wav       & right          & right      & N/A       & N/A        & N/A    & N/A       \\
Right\_0\_pert\_pen.wav & five           & right       & Penalty   & Untargeted & 25.784 & -24.836   \\
Right\_0\_pert\_itr.wav & six            & right      & iterative & Untargeted & 27.580 & -18.416   \\
On\_0\_org.wav          & on             & on         & N/A       & N/A        & N/A    & N/A       \\
On\_0\_pert\_pen.wav    & sheila         & on         & Penalty   & Untargeted & 25.126 & -25.037   \\
On\_0\_pert\_itr.wav    & stop           & on         & iterative & Untargeted & 26.609 & -18.416   \\
Eight\_0\_org.wav       & eight          & eight      & N/A       & N/A        & N/A    & N/A       \\
Eight\_0\_pert\_pen.wav & sheila         & eight      & penalty   & Untargeted & 25.921 & -24.822   \\
Eight\_0\_pert\_itr.wav & six            & eight      & iterative & Untargeted & 27.764 & -18.416   \\
Two\_0\_org.wav         & two            & two        & N/A       & N/A        & N/A    & N/A       \\
Two\_0\_pert\_pen.wav   & sheila         & two        & penalty   & targeted   & 26.627 & -24.813   \\
Two\_0\_pert\_itr.wav   & sheila         & two        & iterative & Untargeted & 28.200 & -18.416   \\
No\_1\_org.wav          & no             & no         & N/A       & N/A        & N/A    & N/A       \\
No\_1\_pert\_pen.wav    & sheila         & no         & penalty   & Untargeted & 27.316 & -24.840   \\
No\_1\_pert\_itr.wav    & sheila         & no         & iterative & Untargeted & 28.845 & -18.416   \\
\bottomrule
\end{tabular}
\caption{List of examples of perturbed audio samples, Methodology of crafting the samples, SNR of the perturbed samples, $dB_{\mathbf x}(\mathbf v)$ of the perturbed samples and also detected class of the sample by SpchCMD model and the true class of the samples. The audio files belong to Speech Commands dataset \cite{warden2018speech}. N/A: Not Applicable}
\label{tab:examples_spch}

\end{table*}
\end{center}


\ifCLASSOPTIONcompsoc
  \section*{Acknowledgments}
\else
  \section*{Acknowledgment}
\fi

We thank Dr. Rachel Bouserhal for her insightful feedback. This work was funded by the Natural Sciences and Engineering Research Council of Canada (NSERC). This work was also supported by the NVIDIA GPU Grant Program. 

\ifCLASSOPTIONcaptionsoff
  \newpage
\fi



\bibliographystyle{IEEEtran}
\bibliography{bib.bib}

\vfill

\begin{IEEEbiography}[{\includegraphics[width=1in,height=1.25in,clip,keepaspectratio]{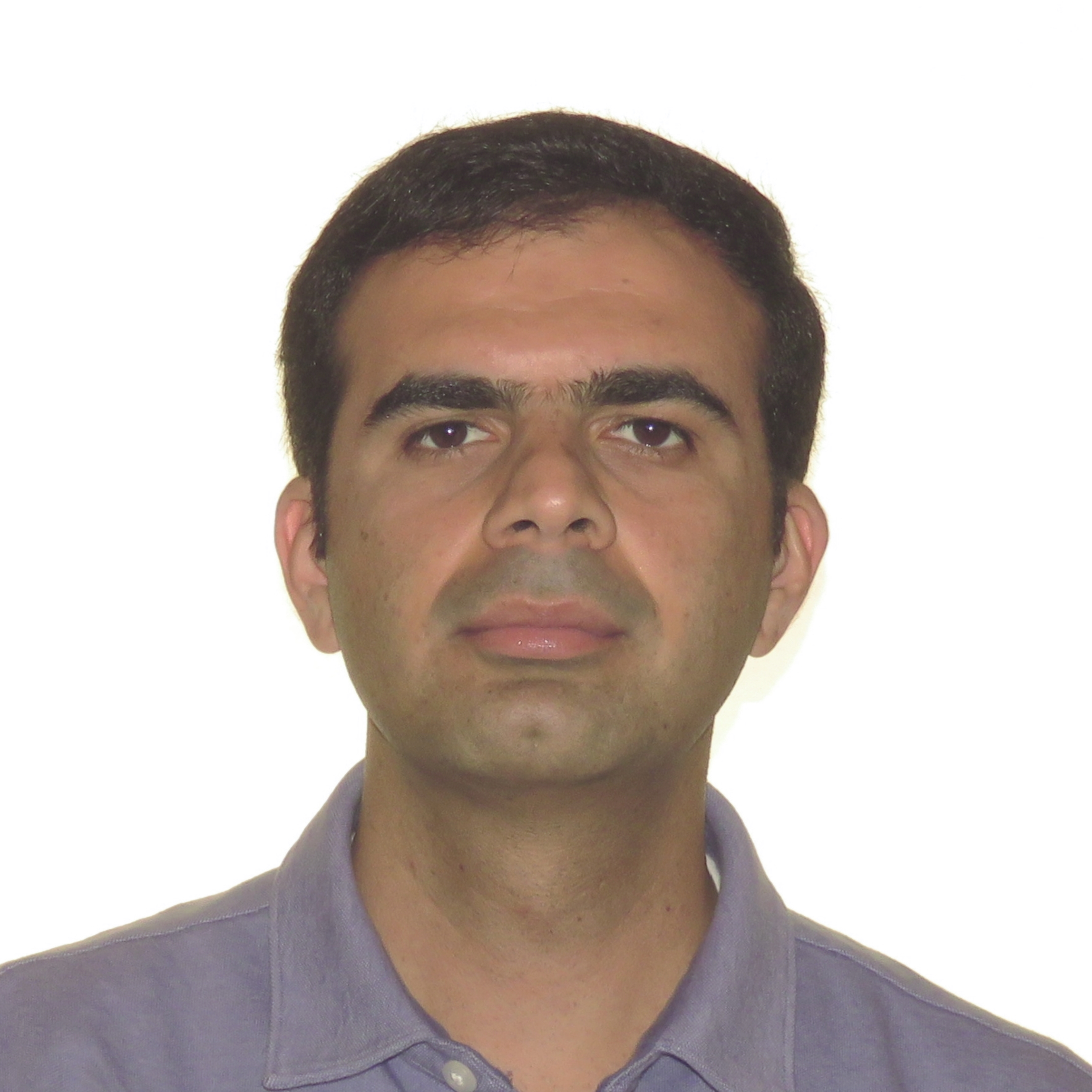}}]{Sajjad Abdoli}
Received his Master's degree in computer engineering from QIAU, Qazvin, Iran in 2017. He is currently a Ph.D candidate at \'Ecole de Technologie Sup\'erieure (\'ETS), Universit\'e du Qu\'ebec, Montreal, QC, Canada. His research interests include audio and speech processing, music information retrieval and developing adversarial attacks on machine learning systems.  
\end{IEEEbiography}
\begin{IEEEbiography}[{\includegraphics[width=1in,height=1.25in,clip,keepaspectratio]{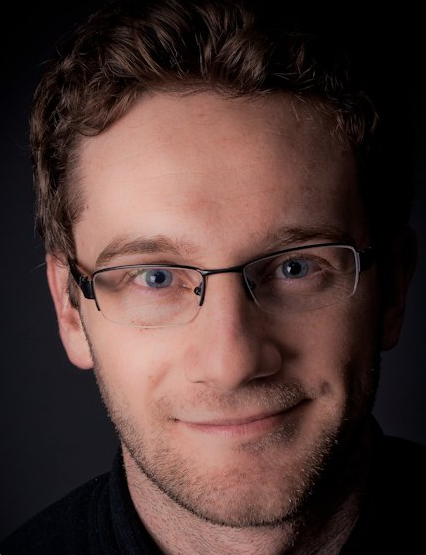}}]{Luiz Gustavo Hafemann} received his B.S. and M.Sc. degrees in Computer Science from the Federal University of Paran\'a, Brazil, in the years of 2008 and 2014, respectively. He received his Ph.D. degree in Systems Engineering in 2019 from the \'Ecole de Technologie Sup\'erieure, Canada. He is currently a researcher at Sportlogiq, applying computer vision models for sports analytics. His interests include meta-learning, adversarial machine learning and group activity recognition.
\end{IEEEbiography}
\begin{IEEEbiography}[{\includegraphics[width=1in,height=1.25in,clip,keepaspectratio]{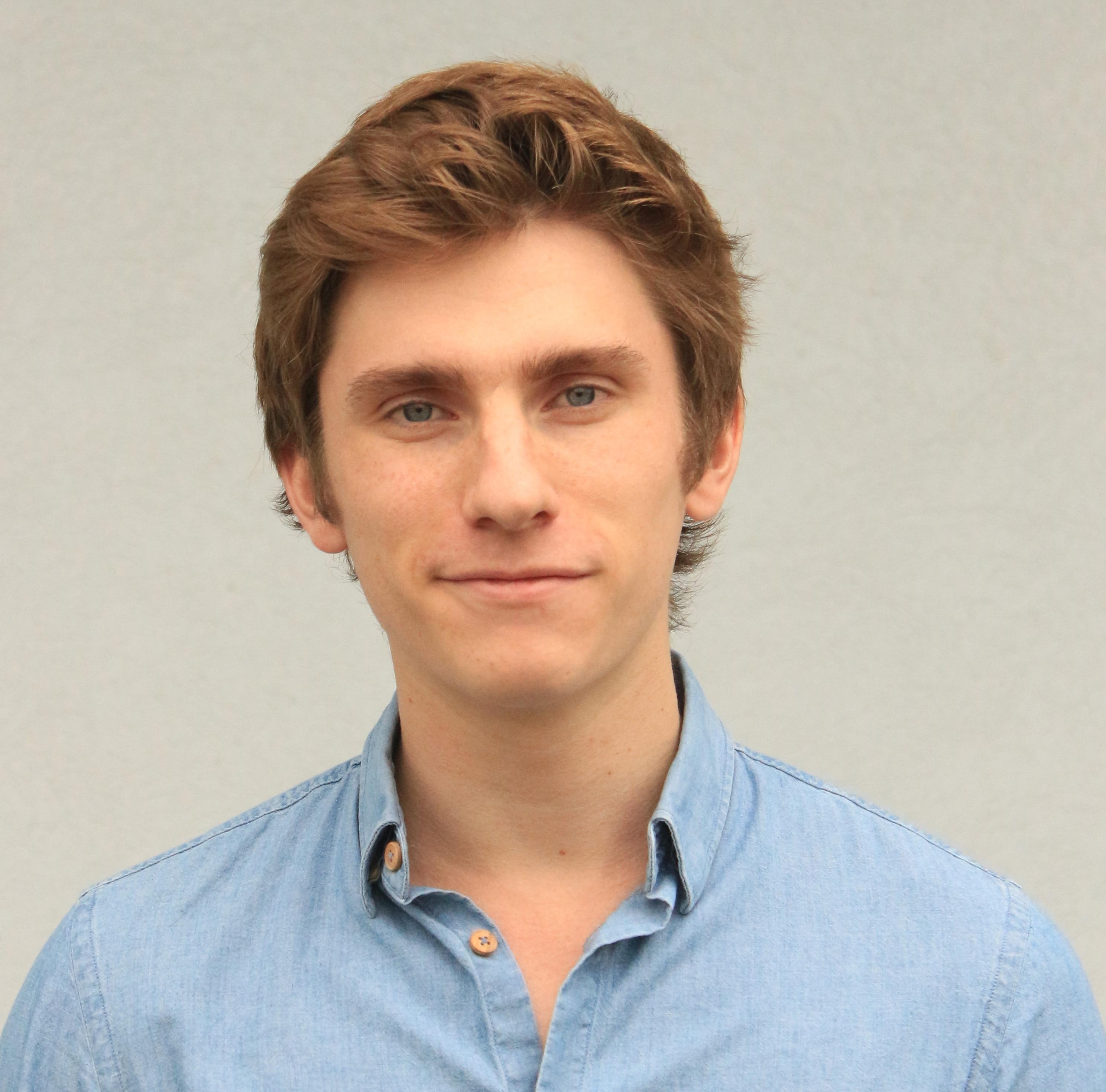}}]{J\'er\^ome Rony} received his M.A.Sc. degree in Systems Engineering in 2019 from the \'Ecole de Technologie Sup\'erieure (\'ETS), Universit\'e du Qu\'ebec, Montreal, QC, Canada. He is currently a Ph.D student at \'ETS whose research interests include computer vision, adversarial examples and robust machine learning.
\end{IEEEbiography}
\begin{IEEEbiography}[{\includegraphics[width=1in,height=1.25in,clip,keepaspectratio]{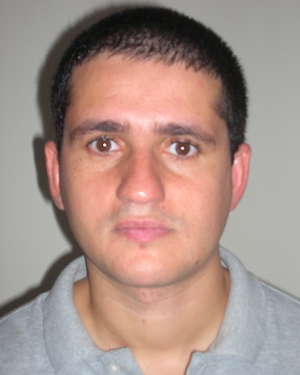}}]{Ismail Ben Ayed}
received the PhD degree (with the highest honor) in computer vision from the Institut National de la Recherche Scientifique (INRS-EMT), Montreal, QC, in 2007. He is currently Associate Professor at the Ecole de Technologie Superieure (ETS), University of Quebec, where he holds a research chair on Artificial
Intelligence in Medical Imaging. Before joining the ETS, he worked for 8 years as a research scientist at GE Healthcare, London, ON, conducting
research in medical image analysis. His research interests are in computer vision, optimization, machine learning and their potential applications in medical image analysis. 
\end{IEEEbiography}
\begin{IEEEbiography}[{\includegraphics[width=1in,height=1.25in,clip,keepaspectratio]{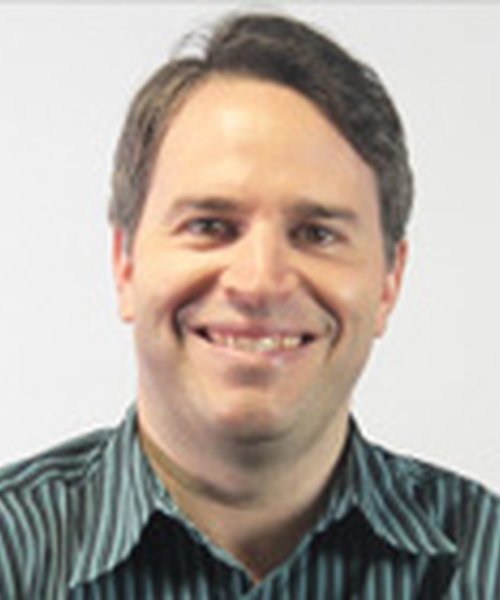}}]{Patrick Cardinal}received the B. Eng. degree in electrical engineering in 2000 from \'{E}cole de Technologie Sup\'{e}rieure (\'{E}TS), M.Sc. from McGill University in 2003 and PhD from \'{E}TS in 2013. From 2000 to 2013, he has been involved in several projects related to speech processing, especially in the development of a closed-captioning system for live television shows based on automatic speech recognition. After his postdoc at MIT, he joined \'{E}TS as a professor. His research interests cover several aspects of speech processing for real life and medical applications.  
\end{IEEEbiography}
\begin{IEEEbiography}[{\includegraphics[width=1.0in,height=1.25in,clip]{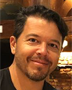}}]
{Alessandro Lameiras Koerich} is an Associate Professor in the Dept. of Software and IT Engineering of the \'{E}cole de Technologie Sup\'{e}rieure (\'{E}TS). He received the B.Eng. degree in electrical engineering from the Federal University of Santa Catarina, Brazil, in 1995, the M.Sc. in electrical engineering from the University of Campinas, Brazil, in 1997, and the Ph.D. in engineering from the \'{E}TS, in 2002. His current research interests include computer vision, machine learning and music information retrieval.
\end{IEEEbiography}




\vfill


\end{document}